\pdfoutput=1

\documentclass[11pt]{article}

\usepackage{acl}

\usepackage[dvipsnames]{xcolor}
\usepackage{times}
\usepackage{latexsym}
\usepackage{graphicx}
\usepackage{textcomp}
\usepackage{colortbl}
\usepackage{xcolor}
\usepackage{amsmath} 
\usepackage{booktabs} 
\usepackage{array}
\usepackage{graphicx} 
\usepackage{multirow}
\usepackage{color}
\usepackage{makecell} 
\usepackage{float}
\usepackage{subcaption}
\usepackage{colortbl}
\usepackage{placeins}
\usepackage[ruled,vlined]{algorithm2e}
\usepackage{comment}
\usepackage{pifont}
\usepackage{url} 
\usepackage{tcolorbox}
\tcbuselibrary{breakable, skins, theorems}

\usepackage[T1]{fontenc}
\usepackage[utf8]{inputenc}
\usepackage{microtype}
\usepackage{inconsolata}
\usepackage{graphicx}

\usepackage{diagbox}
\usepackage{xcolor}
\usepackage{listings}
\usepackage{tcolorbox}

\title{CapGeo-Bench: Decoupling Visual Perception from Reasoning and Evaluating Geometric Understanding}

\author{\textbf{Yuying Li}$^{1}$\thanks{Equal Contribution.}, \ 
\textbf{Siyi Qian}$^{2*}$, \ 
\textbf{Hao Liang}$^{2*}$, \ 
\textbf{Leqi Zheng}$^{1}$, \ 
\textbf{Ruichuan An}$^{2}$, \\
\textbf{Linzhuang Sun}$^{3}$, \
\textbf{Jiajun Zhang}$^{4}$, \
\textbf{Wentao Zhang}$^{2}$\thanks{Corresponding Author.}\\
\textsuperscript{1}Tsinghua University,
\textsuperscript{2}Peking University,
\textsuperscript{3}University of Chinese Academy of Sciences,\\
\textsuperscript{4}University of Science and Technology of China\\
\tt\small liyuying25@mails.tsinghua.edu.cn \quad wentao.zhang@pku.edu.cn\\
}

\begin{document}
\maketitle
\begin{abstract}
While Multimodal Large Language Models (MLLMs) have achieved remarkable success in difficult purely textual mathematical reasoning tasks, even advanced closed-source models such as GPT-o3 still struggle with geometric problems. This discrepancy motivates us to investigate the root cause: is the bottleneck of multimodal geometric reasoning rooted in reasoning itself, or in the perception of geometric information from diagrams? To answer this question, we conduct an exploratory experiment and find that providing high-quality captions consistently and substantially boosts performance of MLLMs, empirically validating the visual perception bottleneck in geometric reasoning. However, MLLMs' capabilities in visual geometric perception remain insufficiently evaluated. To fill the gap, we propose \textbf{CapGeo-Bench}, a benchmark of 4,641 high-quality figure–caption pairs equipped with a fine-grained keypoint-based evaluation metric. Specifically, CapGeo-Bench encompasses three geometric classes and is devided into four difficulty levels according to the complexity of geometric primitives. Captions are scored at a fine-grained level by extracting keypoints along three dimensions-geometric elements, spatial relationships, and numerical attributes. CapGeo-Bench provides a reliable instrument and guideline for evaluating and advancing MLLMs' geometric perception capabilities.
Our code and data are publicly available at: \url{https://github.com/YuYingLi0/CapGeo}.

\end{abstract}

\section{Introduction}\label{sec:intro}

\begin{figure*}[htbp]
    \centering
    \includegraphics[width=1.0\linewidth]{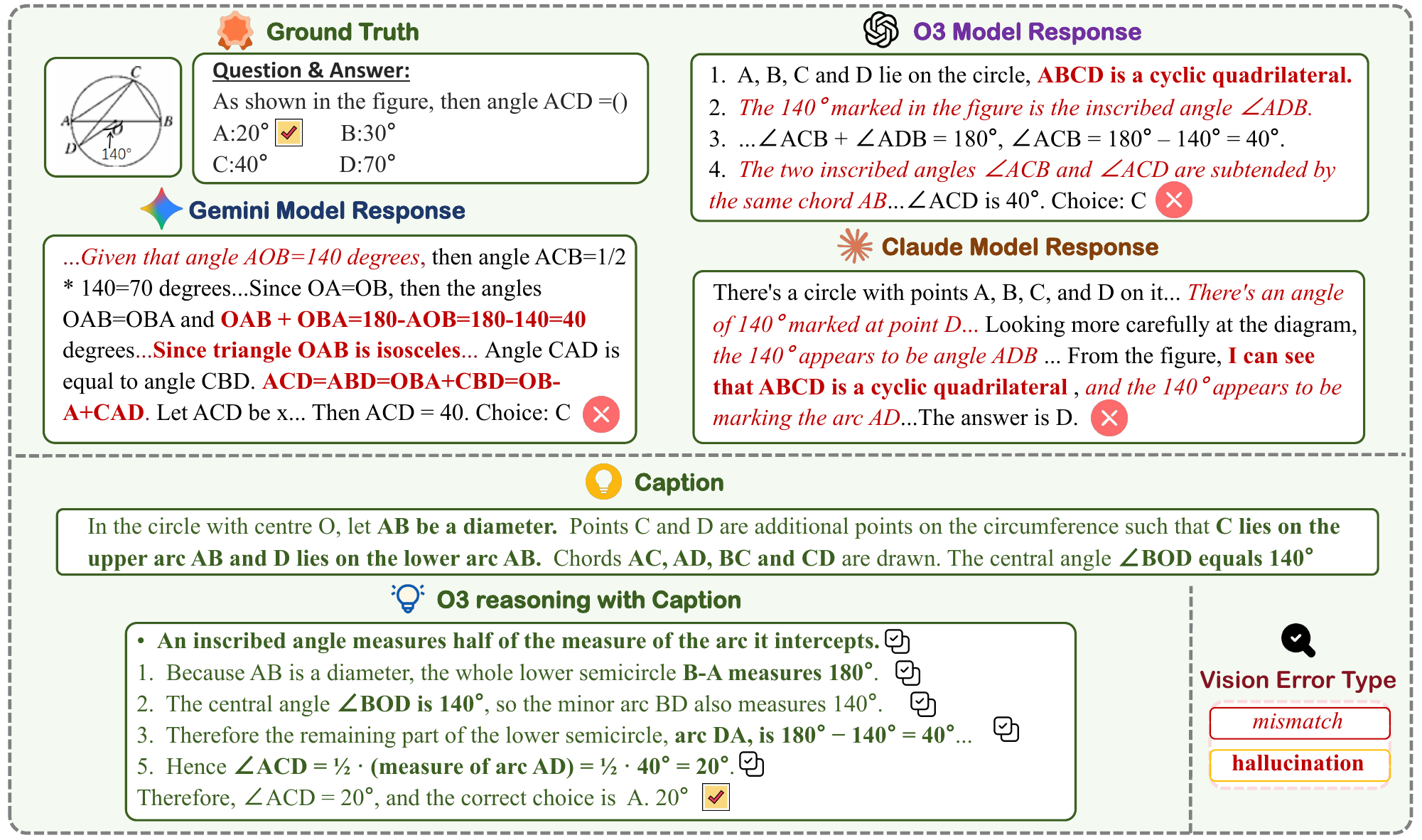}
    \caption{Bad Cases in Mathverse. The upper shows the incorrect results of MLLMs. \textit{Mismatch} means generated relationship exists, but the relationship subjects mismatch. After caption assistance, GPT-o3 reasons correctly.}
    \label{fig: Bad_case}
    \vspace{-2mm}
\end{figure*}

Although LLMs and MLLMs have achieved promising results across diverse tasks~\cite{bai2025qwen2, yao2024minicpm,10.1145/3774904.3792075}, multimodal reasoning remains a central challenge. Geometric benchmarks such as MathVerse~\cite{zhang2024mathverse}, MathVista~\cite{lu2023mathvista}, and GeoQA~\cite{chen2021geoqa} further underscore this limitation: even GPT-o3 and Gemini2.5-Pro~\cite{gemini2025report} often struggle to solve geometry-related problems. In contrast, they demonstrate much greater stability and accuracy in text-only reasoning tasks such as the International Mathematical Olympiad (IMO)~\cite{huang2025gemini} and International Physics Olympiad (IPhO)~\cite{qiu2025physics}.
This stark discrepancy raises a critical question- \textit{What exactly are the bottlenecks that constrain the multimodal geometric reasoning of the existing MLLMs?} 

As shown in Figure~\ref{fig: Bad_case}, MLLMs fail to obtain clear geometric conditions from visual input, producing hallucinations or mismatches that derail subsequent reasoning. In contrast, when accurate caption is supplied, GPT-o3 reasons correctly. Inspired by this observation and relevant research~\cite{wei2024slow, chen2025geopqa}, we think that imperfect visual perception—rather than reasoning itself—is the dominant factor constraining geometric reasoning under current difficulty levels.

To verify this hypothesis, we design \textbf{CapGeo}, a caption-assisted reasoning setup that decouples perception from reasoning. Specifically, we systematically compare model performance under three input modalities: Image-only, Caption-only, and Caption+Image. As shown in Table~\ref{tab:capgeo_preview}, reasoning models equipped with GPT-o3-generated captions yields dramatic and consistent improvements over the Image-only baseline on MathVerse. We further conduct two ablation studies to rule out alternative explanations like textual prior knowledge and any specific template design. Together, these results empirically confirm that visual perception, rather than reasoning, is the bottleneck of multimodal geometric reasoning for current MLLMs.

This finding also aligns with a recent trend in multimodal reinforcement learning, where researchers explicitly inject perception-level supervision or description-based reward signals into the training~\cite{xing2025caprl,chen2025perception,xiao2025perception,xia2025visionary}.
However, \textit{how systematically measure the quality of geometric captions and geometric perception capability of MLLMs} is still a blank area. Existing benchmarks and metric for general image captioning fall short for geometry, where descriptions must faithfully encode elements, spatial relations, and numerical constraints. To fill this gap, we introduce \textbf{CapGeo-Bench}, a benchmark of 4{,}641 high-quality geometric figure–caption pairs spanning three geometric classes and four difficulty levels. We further propose a keypoint-based fine-grained evaluation framework that scores captions along three dimensions—elements, spatial relationships, and numerical attributes—providing a principled metric for diagnosing MLLMs' geometric perception. Comprehensive experiments on CapGeo-Bench reveal that current MLLMs still struggle with geometric captioning, particularly in the numerical dimension. Moreover, a joint analysis shows that CapGeo-Bench scores are strongly correlated with downstream reasoning performance, confirming its reliability as a perception evaluation tool.

Our contributions are summarized as follows:
\begin{itemize}
\item We conduct an \textbf{exploratory experiment via CapGeo}, a caption-assisted reasoning setup that decouples perception from reasoning. Comprehensive comparisons across three input modalities verify that visual perception, rather than reasoning, is the bottleneck of multimodal geometric reasoning.

\item Motivated by this finding, we release \textbf{CapGeo-Bench}, a high-quality benchmark of 4{,}641 geometric figure–caption pairs covering a broad spectrum of categories and difficulty levels, dedicated to evaluating MLLMs' geometric information extraction capability.

\item We propose a novel \textbf{keypoint-by-keypoint geometric caption evaluation method} that scores captions along three fine-grained dimensions validated by multiple mathematics experts. This method provides a principled, interpretable, and reproducible metric for diagnosing MLLMs' geometric perception. Scores on CapGeo-Bench show a strong correlation with downstream reasoning performance, further confirming the reliability of our evaluation framework.
\end{itemize}

\section{Related Work}
\begin{figure*}
    \centering
    \includegraphics[width=1.0\linewidth]{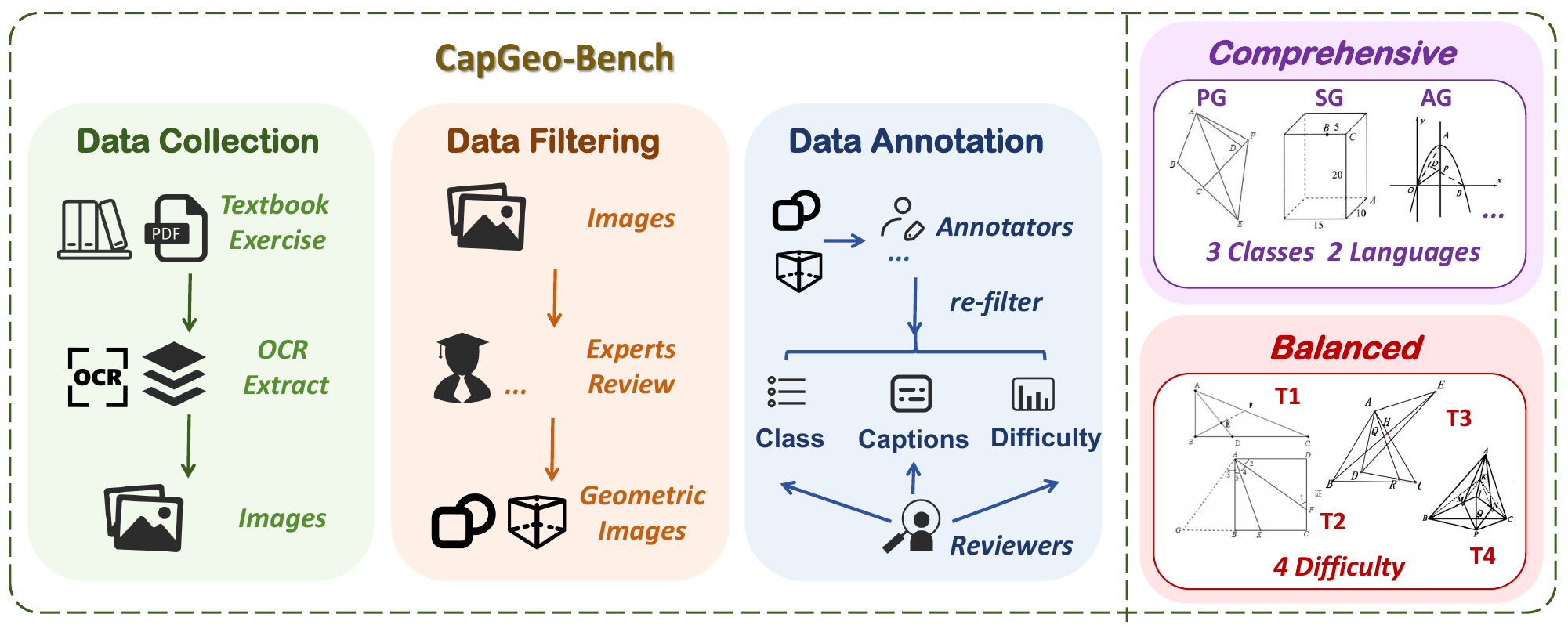}
    \caption{Overview of CapGeo-Bench. AG: Analytic Geometry, PG: Plane Geometry, SG: Solid Geometry}
    \label{fig:capbench}
    \vspace{-4mm}
\end{figure*}
\subsection{MLLMs Reasoning}
Recently, The performance of MLLMs shone brightly.
High-performance MLLMs include OpenAI's o3~\cite{openai2025o3}, Qwen2.5-VL series~\cite{qwen25vl}, and LLaVA series~\cite{llava}.

Current paradigms for enhancing multimodal reasoning capabilities of MLLMs encompass~\cite{mmsurvey}: (1) Multimodal Chain-of-Thought (MCoT), which employs prompt engineering to guide LLs through step-by-step reasoning processes~\cite{mcot1,mcot2, mcot3, mcot4}; (2) Multimodal-o1 models~\cite{mo1-1, mo1-2, mo1-3}; 
(3) Multimodal-R1 models~\cite{mr1-1, mr1-2}.

\subsection{Geometric Reasoning}
The research on multimodal geometric reasoning emerge constantly. G-LLAVA~\cite{gao2023g} trained the G-LLaVA model using the geometric synthesis dataset Geo170K; and MMGeoLM~\cite{mmgeolm} constructed positive and negative sample pairs for contrastive learning training. Benchmarks such as MathVista~\cite{lu2023mathvista} and MathVision~\cite{mathv} assess the geometric reasoning capabilities of MLLMs in multiple geometric domains. Some research explores the pseudo-visual phenomena in existing MLLMs, such as MathVerse~\cite{zhang2024mathverse}.

Compared with studies like Describe-then-Reason~\cite{jia2024describe} framework, CapGeo is designed for exploratory experiment. It decouples perception from reasoning, providing conditions for a deep analysis of the performance differences under different input modalities to verify our view.


\subsection{Image Captioning}
Image captioning is a fundamental direction in Computer Vision. Broadly speaking, there are two main approaches: (i) direct caption generation. In the era of MLLMs, enhancing the captioning capabilities of MLLMs remains a significant focus for many researchers~\cite{caption1,caption2,caption3}. Meanwhile, the evaluation of image captions itself has also become an important research direction~\cite{captioneval, captioneval2}; (ii) OCR-based methods. For MLLMs, OCR focuses on extracting pre-existing textual content from images. This capability not only emphasizes information extraction performance~\cite{yang2021textvqa, mathew2021docvqa, fu2024ocrbench, liu2024ocrbench}, but also highlights the reasoning ability required in real-world scenarios~\cite{reasoningocr, ocrreasoning}.

Nevertheless, most studies target general natural images. Some geometric datasets include captions, but they lack evaluation~\cite{peng2024multimath}. To address this gap, we propose \textit{CapGeo-Bench}.

\begin{figure*}[htbp]
\centering
\begin{minipage}[h]{0.33\textwidth}
    \centering
    \resizebox{0.9\linewidth}{!}{
    \begin{tabular}{p{3.5cm}p{0.8cm}}
    \toprule
    \textbf{Statistic} & \textbf{Value} \\
    \midrule
    Total images & 4,641 \\
    \midrule
    \quad - Difficulty 1 & 1,816 \\
    \quad - Difficulty 2 & 2,133 \\
    \quad - Difficulty 3 & 495 \\
    \quad - Difficulty 4 & 197 \\
    \midrule
    \quad - Plane Geometry & 4,023 \\
    \quad - Analytic Geometry & 517 \\
    \quad - Solid Geometry & 101 \\
    \midrule
    Maximum word & 586 \\
    Average word & 61.4 \\
    \bottomrule
    \end{tabular}
    }
    \captionof{table}{CapGeo-Bench Statistics}
    \label{tab:dataset_statistics}
\end{minipage}
\hfill
\begin{minipage}[h]{0.66\textwidth}
    \centering    
    \includegraphics[width=\linewidth]{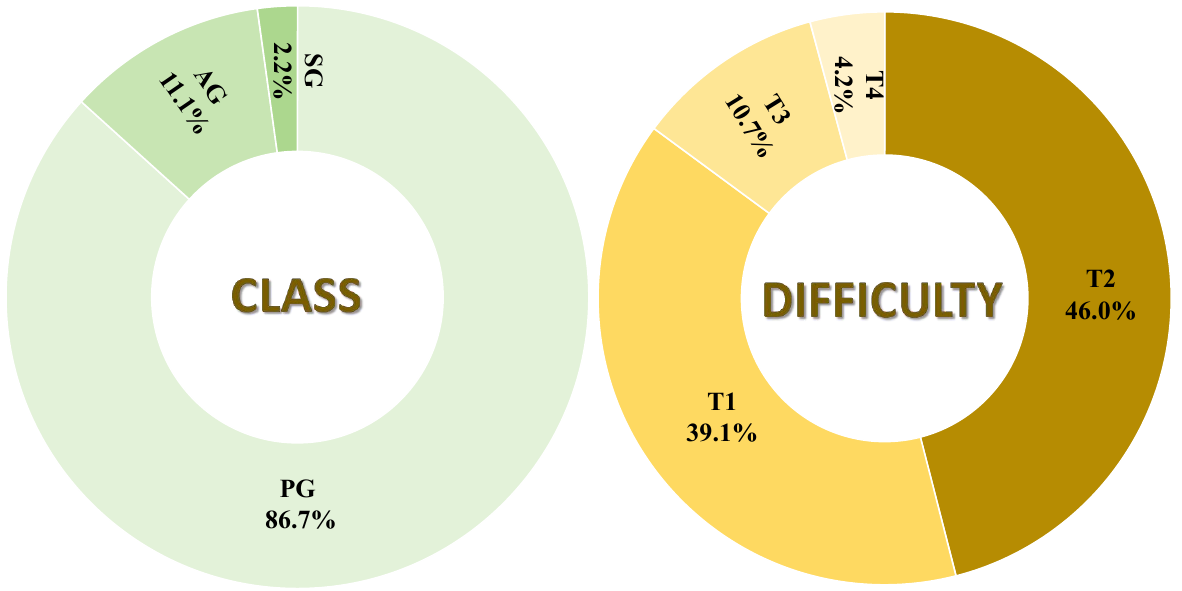}
    \captionof{figure}{CapGeo-Bench Data Statistics Chart}
    \label{fig:chart}
\end{minipage}

\end{figure*}

\section{CapGeo: An Exploratory Experiment on the Perception Bottleneck}




\subsection{Experimental Design}
CapGeo consists of two stages: a captioning stage that converts geometric figures into textual captions, and a reasoning stage that consumes the caption (and optionally the original figure) for solving.

\textbf{Captioning stage.} We instruct the captioner to describe all visual information in the figure using formal geometric language. see Figure~\ref{fig:caption} for the prompt template.

\textbf{Reasoning stage.} Given a question $Q$, the original image $I$, and the generated caption $C=\text{Caption}(I)$, the reasoning model outputs an answer
$$
A = \text{MLLM}(Q, I, C).
$$
By selectively masking $I$ or $C$, we obtain three input modalities:
(i) \textbf{Image only} ($Img$): the conventional vision-based setting;
(ii) \textbf{Caption only} ($Cap$): pure text-based reasoning;
(iii) \textbf{Caption + Image} ($Cap{+}Img$): joint multimodal input.

This controlled design allows us to disentangle the contributions of perception and reasoning.

\begin{figure*}[htbp]
    \centering
    \includegraphics[width=1.0\linewidth]{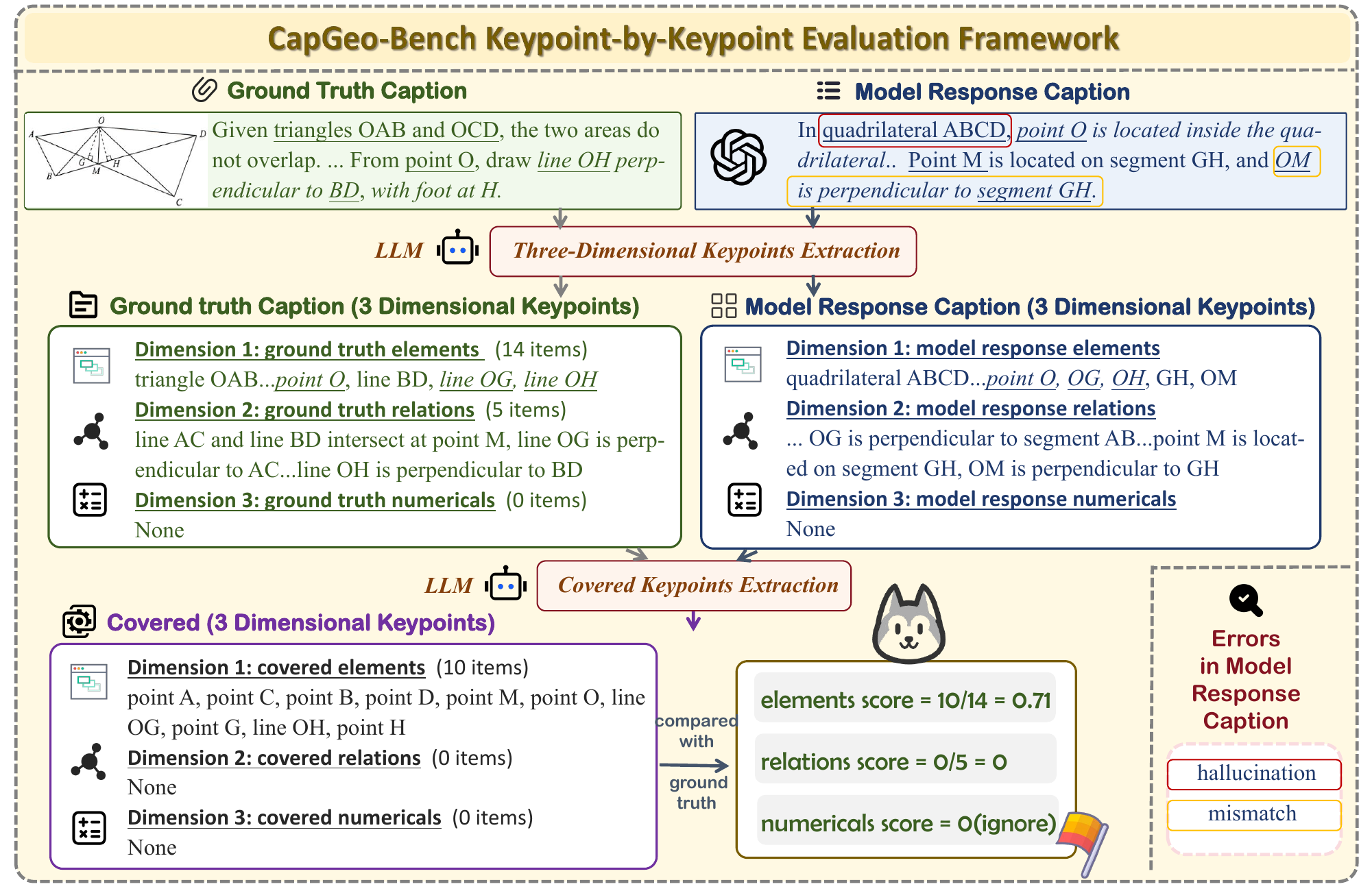}
    \caption{Overview of CapGeo-Bench Evaluation. Covered items in keypoints are marked by italics and underlining.}
    \label{fig:eval}
    \vspace{-4mm}
\end{figure*}

\subsection{Experimental setup and result}
\textbf{Experimental setup.} We employ four captioning models (GPT-o3, GPT-4o, Qwen2.5-VL-72B-Instruct, Qwen2.5-VL-7B-Instruct~\cite{qwen25vl}) and eight reasoning models spanning both MLLMs (Claude-Opus-4~\cite{anthropic2025claude}, Gemini-2.5-pro~\cite{gemini2025report}, GPT-o3, Qwen2.5-VL-72B-Instruct, Qwen2.5-VL-7B-Instruct) and LLMs (DeepSeek-R1~\cite{guo2025deepseek}, Qwen2.5-72B-Instruct, and Qwen2.5-7B-Instruct~\cite{bai2025qwen2}). We evaluate on MathVerse~\cite{zhang2024mathverse} (Vision-Only and Vision-Intensive), MathVista~\cite{lu2023mathvista} (geometry subset), and GeoQA~\cite{chen2021geoqa} (200 sampled problems). All models use temperature 0.0 for deterministic output.



\textbf{Verification.} Table~\ref{tab:capgeo_preview} presents results on MathVerse using GPT-o3 as the captioner, with full results across all four captioners in Appendix~\ref{Appendix.fullcap} and additional results on MathVista and GeoQA in Appendix~\ref{Appendix.mvgeoqa}. All consistently show that caption-assisted reasoning substantially outperforms image-only baselines. For instance, on MathVerse (Vision-Only), Qwen2.5-VL-72B-Instruct surges from 8.6\% to 66.4\%, and Claude-Opus-4 from 44.8\% to 73.0\%. Crucially, even text-only LLMs (e.g., DeepSeek-R1, Qwen2.5-72B-Instruct) reach performance comparable to MLLMs once captions are supplied, indicating that captions carry sufficient geometric information for reasoning.

\textbf{Ablation 1: Ruling out textual priors.} A potential alternative explanation is that captions merely activate the model's textual prior knowledge rather than truly conveying visual information. To rule this out, we compare CapGeo's \textit{Caption-only} setting against MathVerse's native \textit{Text-only} setting, which provides only the textual question (without image or caption) and thus precisely isolates the pure textual prior. As shown in Table~\ref{tab:text_prior}, Caption-only consistently and substantially outperforms Text-only—e.g., Qwen2.5-VL-72B-Instruct improves from 62.2\% to 66.1\% (Vision-Only) and 74.9\% (Vision-Intensive). Since the Text-only setting provides neither image nor caption, the model can only rely on textual prior knowledge. The consistent and substantial gap between Caption-only and Text-only directly quantifies the contribution of extracted visual information, ruling out textual priors as the source of improvement.

\textbf{Ablation 2: Ruling out template exploitation.} Another concern is whether the gains depend on a specific captioning format. To address this, we replace our natural-style captions with a rigid JSON schema that explicitly enumerates geometric elements, relationships, and numerical information (Appendix~\ref{Appendix.style}). We observe consistent performance degradation across all reasoning models, indicating that overly formalized representations disrupt the semantic continuity that LLMs naturally exploit. This confirms that CapGeo's improvements arise from accurate visual content extraction rather than from exploiting any particular template design.

Together, these results robustly confirm our hypothesis: \textbf{visual perception, rather than reasoning, is the bottleneck of multimodal geometric reasoning.}

\section{CapGeo-Bench: Grounding Geometric Captioning Ability}
The exploratory experiment confirms that geometric perception, manifested concretely as captioning quality, is the dominant factor governing downstream reasoning accuracy. This finding motivates a natural follow-up question - how can we systematically measure the geometric captioning—i.e., perception—capability of MLLMs? Existing image-captioning benchmarks are inadequate for geometry, which demands faithful encoding of elements, spatial relations, and numerical constraints. To fill this gap, we introduce \textbf{CapGeo-Bench}.

\subsection{Benchmark Construction}
\textbf{Overview.} To bridge the gaps in evaluating the visual geometric information extraction capabilities of MLLMs, we designs CapGeo-Bench. As shown in Figure~\ref{fig:chart} and Table~\ref{tab:dataset_statistics}, CapGeo-Bench is a benchmark comprising high-quality 4641 geometric images with corresponding bilingual (Chinese and English) captions, covering 3 geometric classes and 4 difficulty levels. The entire benchmark design process is shown in Figure~\ref{fig:capbench}.

\begin{table*}[t]
\renewcommand{\arraystretch}{1.1}
\small
\centering
\resizebox{1.0\linewidth}{!}{
\begin{tabular}{l|ccccc|ccc}
\toprule
\multirow{2}{*}{\textbf{Models}}
& \multicolumn{5}{c|}{\textbf{MLLMs}} & \multicolumn{3}{c}{\textbf{LLMs}} \\
\cmidrule(lr){2-6}\cmidrule(lr){7-9}
& \textbf{GPT-o3} & \textbf{Gemini-2.5} & \textbf{Claude-4} & \textbf{Qwen-VL-72B} & \textbf{Qwen-VL-7B} & \textbf{DS-R1} & \textbf{Qwen-72B} & \textbf{Qwen-7B} \\
\midrule
\multicolumn{9}{c}{\textbf{Vision Only}} \\
\midrule
\textbf{Img}     & 74.6 & 66.4 & 44.8 & 8.6  & 7.1  & --   & --   & -- \\
\textbf{Cap}     & 60.3 & 60.3 & 59.4 & 66.1 & \textbf{61.9} & \textbf{68.2} & \textbf{66.1} & \textbf{63.3} \\
\textbf{Cap+Img} & \textbf{75.8} & \textbf{73.6} & \textbf{73.0} & \textbf{66.4} & 60.0 & -- & -- & -- \\
\midrule
\multicolumn{9}{c}{\textbf{Vision Intensive}} \\
\midrule
\textbf{Img}     & 73.0 & 79.6 & 61.8 & 58.0 & 43.3 & --   & --   & -- \\
\textbf{Cap}     & \textbf{82.5} & 83.0 & 79.3 & \textbf{74.9} & \textbf{63.2} & \textbf{71.9} & \textbf{73.2} & \textbf{65.9} \\
\textbf{Cap+Img} & 78.3 & \textbf{85.4} & \textbf{85.5} & 72.6 & 60.3 & -- & -- & -- \\
\bottomrule
\end{tabular}}
\caption{Performance on \textbf{MathVerse} with GPT-o3 as captioner. Full results in Appendix~\ref{Appendix.fullcap}.}
\label{tab:capgeo_preview}

\end{table*}

\begin{table*}[htbp]
\centering
\small
\begin{tabular}{
    p{1.9cm}|
    p{1.5cm}|
    *{3}{>{\centering\arraybackslash}p{0.8cm}}|
    *{4}{>{\centering\arraybackslash}p{0.8cm}}|
    >{\centering\arraybackslash}p{1.0cm}|
    >{\centering\arraybackslash}p{1.0cm}
}
\toprule
\multirow{2}{*}{\textbf{Model}} & \multirow{2}{*}{\textbf{\makecell{Score\\Dimension}}} & \multicolumn{3}{c|}{\textbf{Class}} & \multicolumn{4}{c|}{\textbf{Difficulty}} & \multirow{2}{*}{\textbf{Overall}} & \multirow{2}{*}{\textbf{Avg}} \\
\cmidrule(lr){3-5} \cmidrule(lr){6-9}
 & & \textbf{AG} & \textbf{PG} & \textbf{SG} & \textbf{T1} & \textbf{T2} & \textbf{T3} & \textbf{T4} & & \\
\midrule
\multirow{3}{*}{\textbf{GPT-o3}}
 & Element   & 69.0 & 55.1 & 49.5 & 57.4 & 54.8 & 34.6 & 32.1 & 56.0 & \multirow{3}{*}{38.4} \\
 & Relation  & 41.0 & 38.9 & 26.8 & 41.1 & 35.7 & 20.7 & 20.3 & 38.9 &  \\
 & Numerical & 7.8  & 20.8 & 52.8 & 24.1 & 14.2 & 13.3 & 0.0  & 20.4 &  \\
\midrule
\multirow{3}{*}{\textbf{GPT-4o}}
 & Element   & 64.0 & 53.2 & 49.4 & 54.6 & 54.6 & 51.6 & 50.8 & 54.2 & \multirow{3}{*}{30.4} \\
 & Relation  & 30.9 & 26.6 & 15.5 & 32.3 & 24.3 & 18.2 & 17.8 & 26.8 &  \\
 & Numerical & 19.2 & 9.3  & 24.5 & 12.2 & 10.1 & 3.5  & 3.6  & 10.2 &  \\
\midrule
\multirow{3}{*}{\textbf{\makecell{Qwen2.5-VL\\ -72B-Instruct}}}
 & Element   & 64.6 & 53.0 & 48.2 & 54.7 & 54.6 & 50.2 & 50.4 & 54.1 & \multirow{3}{*}{30.8} \\
 & Relation  & 33.2 & 28.5 & 20.0 & 33.4 & 26.9 & 20.8 & 18.4 & 28.9 &  \\
 & Numerical & 20.1 & 8.3  & 21.4 & 12.4 & 8.1  & 5.6  & 0.0  & 9.4  &  \\
\midrule
\multirow{3}{*}{\textbf{\makecell{Qwen2.5-VL\\ -7B-Instruct}}}
 & Element   & 58.1 & 44.6 & 31.9 & 47.0 & 46.3 & 45.8 & 44.3 & 46.5 & \multirow{3}{*}{24.2} \\
 & Relation  & 26.6 & 17.5 & 2.2  & 23.4 & 16.1 & 11.0 & 11.2 & 18.7 &  \\
 & Numerical & 17.7 & 5.9  & 42.9 & 8.7  & 7.5  & 2.6  & 0.0  & 7.4  &  \\
\bottomrule
\end{tabular}
\caption{Captioning Model Performance on \textbf{CapGeo-Bench} grouped by Class and Difficulty.}
\label{tab:model_comparison}

\end{table*}

\textbf{Data Collection.} To minimize potential contamination from open-source data, we deliberately avoided using publicly available datasets. Instead, we curated a collection of geometric problems from K-12 textbooks in China, encompassing classic topics such as triangles, parallelograms, and circles, with difficulty levels ranging from middle school exercises to competition problems. The entire collection process was conducted in strict compliance with copyright and licensing regulations. We then employed the open-source detection tool DocLayout-YOLO~\cite{zhao2024doclayout} to automatically extract geometric diagrams and manually filtered out non-geometric figures. In total, we obtained 5,000 high-quality geometric images.

\textbf{Caption Annotation.} To guarantee the accuracy and quality of captions, we recruited annotators with STEM backgrounds from Peking University, one of the top universities in China, to perform meticulous manual annotation. All annotators underwent rigorous assessments of their mathematical and geometric proficiency, followed by multiple rounds of training and pre-annotation exercises to ensure compliance with the required standards. During annotation, mathematics experts designed a unified annotation instruction, including geometric image re-filtering rules and geometric caption annotation principles. The complete procedures of the whole workflow, together with detailed instructions, are provided in Appendix~\ref{Appendix.instruction}.

A review team composed of three professional reviewers conducts strict cross-checks on all data annotation results. For annotations that fail the review, they will be re-annotated and re-evaluated. See Appendix~\ref{Appendix.instruction} for details. We also adhere to labour ethics during the whole annotation process, ensuring all annotators were well-paid, with each receiving 10 dollars per geometric figure annotation. In the end, we obtained 4,641 high-quality, contamination-free geometric image–caption pairs.

\subsection{Keypoint-by-Keypoint Geometric Caption Quality Evaluation}\label{sec:keypoint_matching}
Although there are some metrics for evaluating image captions quality~\cite{cider, anderson2016spice}, due to the peculiarities of geometric captions,
there is still no scientific, quantitative and fine-grained metric for assessing geometric captions quality generated by MLLMs. 

Geometric figures are composed of basic geometric elements through clear relationships. This structural feature determines that geometric captions can be systematically deconstructed into keypoints set. This cognition is highly consistent with the core idea of Geometric Formal Language~\cite{lu2021inter}. Based on this, we design a keypoint-by-keypoint geometric caption evaluation method with three dimensional evaluation metrics and corresponding evaluation framework.

\textbf{Evaluation Metrics.} 
The abstract combination of geometric points, lines, and letters conveys rich elemental, spatial, and numerical information. 
Thus, for MLLMs’ captioning capability, we focus on the following three dimensional metrics: \\
(1) Elements: evaluate whether MLLMs can comprehensively identify all elements and their corresponding letter identifiers. The elements include basic shapes, lines, points, coordinate axes, etc. In Figure~\ref{fig:eval}, Elements in Captions are underlined. \\
(2) Spatial Relations: evaluate whether MLLMs can comprehensively identify all spatial relations and their corresponding subjects. For specific details, See in Appendix~\ref{Appendix.Spatial}. In Figure~\ref{fig:eval}, Spatial Relations in Captions are in italics.\\
(3) Numerical Relations: evaluate whether the MLLMs can comprehensively identify all the numerical relationships and their corresponding subjects. Like length, angle, etc.

\begin{table}[h]
\centering
\small
\begin{tabular}{lccc}
\toprule
\multirow{2}{*}{\textbf{Reasoning Model}} & \multirow{2}{*}{\textbf{\makecell{Text\\Only}}} & \multicolumn{2}{c}{\textbf{Cap Only (GPT-o3)}}\\
\cmidrule(lr){3-4}
& & \textbf{V-Only} & \textbf{V-Intensive}\\
\midrule
Qwen2.5-VL-7B  & 51.3 & 61.9 & 63.2\\
Qwen2.5-VL-72B & 62.2 & 66.1 & 74.9\\
\bottomrule
\end{tabular}
\caption{Caption-only vs.\ Text-only on MathVerse.}
\label{tab:text_prior}
\end{table}

\textbf{Evaluation Framework.} As shown in Figure ~\ref{fig:eval}, we decompose the evaluation framework into three major steps: 
(1) Extract geometric elements keypoints set \( E\), spatial relation keypoints set \( R\), numerical relation keypoints set \( N\) from both model response captions \( T_c \) and ground truth \( T_g \) by \( LLM \). This can be expressed by the formula as
\[
E,\ R,\ N = \text{LLM}_{\text{extract}}(T)
\]

See Figure~\ref{fig:prompt1} and Figure~\ref{fig:prompt2} for prompts for ground truth and model response caption three dimensional keypoints extraction.

(2) Identify covered items between ground truth keypoints and model response caption keypoints. Due to the varied expressions in natural language, performing a direct intersection operation is rough and insufficient. Therefore, we design prompt and employ LLM to perform semantically equivalent covered items extraction between the model response caption and ground truth keypoints set for each dimension. This process identifies semantically equivalent keypoints, which constitute the \(TP\) set for that dimension. This can be expressed by the formula as
\[
TP_E = E_c \cap E_g = \text{LLM}_{\text{match}}(E_c, E_g)
\]
See Figure~\ref{fig:prompt3} for prompt for covered three dimensional keypoints extraction of \( LLM_{\text{match}}\).\\
(3) The score \(S\) for each dimension is calculated as macro F1. For a dimension $k \in \{E, R, N\}$, we first calculate Precision ($P_k$) and Recall ($R_k$):

\[
P_k = \frac{|TP_k|}{|K_c|}, \quad R_k = \frac{|TP_k|}{|K_g|}
\]

The final score $S$ is expressed by the formula as:

\[
S_k = \frac{2 P_k R_k}{P_k + R_k}
\]

 We submitted the complete evaluation process of 200 geometric captions to three domain experts for independent review. The experts confirmed the validity and reliability of the evaluation processes, which verified the rationality of our method.

\begin{table}[h]
\centering
\small
\begin{tabular}{lcc}
\toprule
\textbf{Reasoning Model} & \textbf{Natural} & \textbf{Structured}\\
\midrule
\multicolumn{3}{l}{\textit{MLLMs}}\\
Qwen2.5-VL-7B-Instruct  & \textbf{32.4} & 31.6\\
Qwen2.5-VL-72B-Instruct & \textbf{39.2} & 36.0\\
\midrule
\multicolumn{3}{l}{\textit{LLMs}}\\
Qwen2.5-7B-Instruct  & \textbf{32.6} & 31.9\\
Qwen2.5-72B-Instruct & \textbf{38.2} & 35.5\\
\bottomrule
\end{tabular}
\caption{Ablation on caption style.}
\label{tab:caption_style}
\end{table}

\section{Experiments}
To further evaluate the captioning capabilities of different MLLMs, we perform a systematic evaluation on CapGeo-Bench. We also conduct a joint analysis linking CapGeo-Bench scores with downstream reasoning performance in CapGeo.

\subsection{Experimental Setup}\label{sec:Experimental_Setup}

\paragraph{Models}
In CapGeo-Bench experiments, we selected GPT-4o, GPT-o3, Qwen2.5-VL-72B-Instruct and Qwen2.5-VL-7B-Instruct as evaluation models and Gemini-2.5-pro as the as the model for extracting keypoints and matching in section~\ref{sec:keypoint_matching}.



\paragraph{Settings}
For local model inference, we employed the VLLM framework~\cite{kwon2023efficient}. The model temperature was fixed at 0.0 to minimize randomness and ensure stable reasoning performance.

\subsection{Experimental Analysis}\label{sec:CapGeo_Bench}


\textbf{Geometric captioning remains a significant challenge.} Table~\ref{tab:model_comparison} reports MLLMs' scores across three keypoint dimensions and four difficulty levels. GPT-o3 attains the best overall average (38.4\%), while Qwen2.5-VL-72B-Instruct performs on par with GPT-4o. Several notable patterns emerge: (1) All models perform dramatically worse on the Numerical dimension, often producing mismatches between numerical values and their corresponding geometric elements—even GPT-o3 achieves only 20.4\% on this dimension. We think that spatial relations rely on explicit physical connections, but numerical values are "floating" text near geometric components without explicit pointers, which make it difficult for MLLM. (2) Performance degrades sharply with increasing difficulty across all three dimensions
. (3) Plane Geometry remains especially challenging despite being the most common category, underscoring the limited ability of current MLLMs to extract information from highly abstract and symbolic visual content. 

These results reveal a substantial gap between MLLM capabilities and the reliable geometric perception, establishing CapGeo-Bench as a challenging and discriminative benchmark.

\begin{figure}[htbp]
\centering
\begin{minipage}[h]{0.49\textwidth}
    \centering    
    \includegraphics[width=\linewidth]{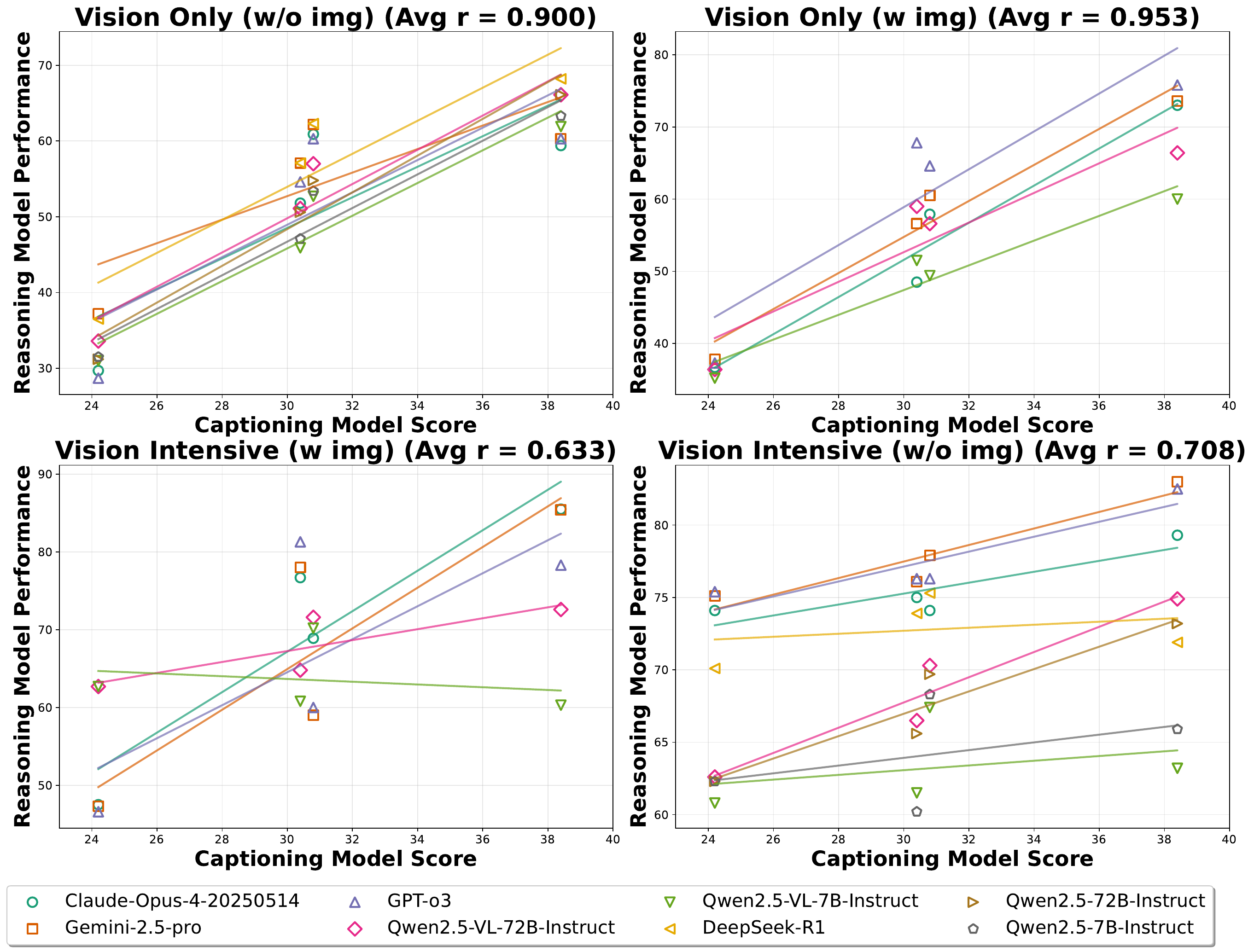}
    \captionof{figure}{Correlation Analysis. w/o img means Cap Only, w img means Cap+Img, r means relative coefficient}
    \label{fig:correlation}
\end{minipage}
\end{figure}

\textbf{Correlation - bridging CapGeo-Bench and downstream reasoning.}
To verify that CapGeo-Bench faithfully reflects perception capability, We conduct a correlation analysis between Three-dimensional average score of the Captioning Model and the performance of the Reasoning Models it assisted in Table~\ref{tab:full_cap}. As shown in Figure~\ref{fig:correlation}, the two are strongly positively correlated—that is, captioners scoring higher on CapGeo-Bench consistently produce captions that lead to better downstream reasoning performance. This correlation (i) reconfirms our hypothesis from a different angle—better perception yields better reasoning—and (ii) Confirms CapGeo-Bench as a reliable evluation tool for geometric understanding.

\textbf{Modality Conflict and Toxic Captions.} Although high-quality captioner improve the performance, the observed abnormal drops need analysis. 

\begin{figure}[htbp]
\centering
\begin{minipage}[h]{0.49\textwidth}
    \centering    
    \includegraphics[width=\linewidth]{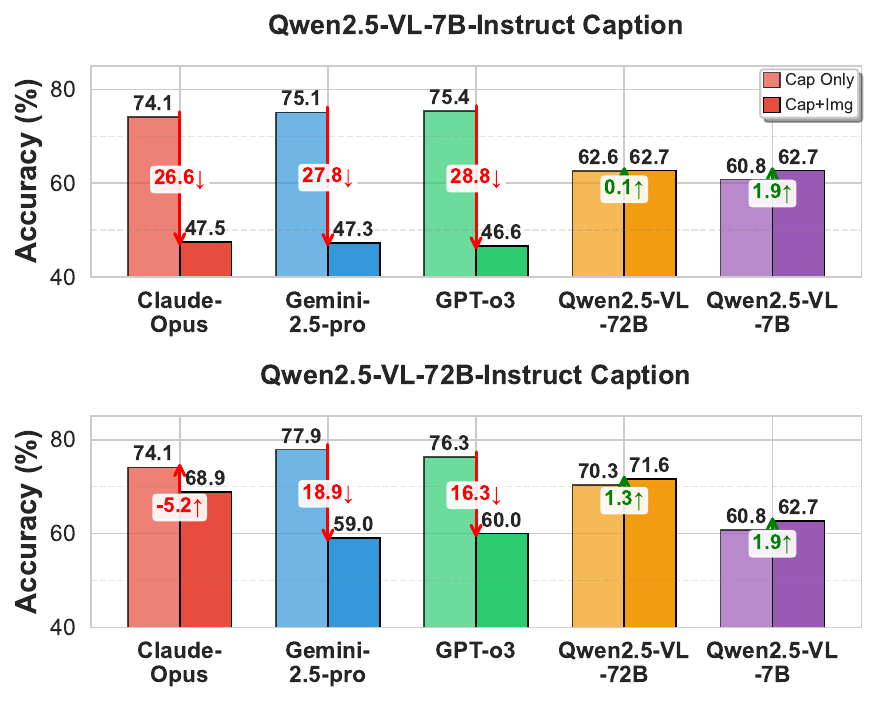}
    \captionof{figure}{Modality Conflict Analysis}
    \label{fig:modality_conflict}

\end{minipage}
\end{figure}

Modality Conflict. As shown in~\ref{fig:modality_conflict}, in Vision Intensive, when using low-quality Captioners, the performance of Cap+Img of the strong MLLMs is lower than Cap Only. For example, for GPT-o3 assisted by captioner Qwen2.5-VL-7B-Instruction, after adding the original image, the accuracy rate drops sharply from 75.4\% to 46.6\%; similarly, on MathVista, Claude-Opus's performance unexpectedly drops from 73.1\% to 65.7\% when assisted by GPT-4o. We think that when there is a contradiction between the caption and the image, MLLMs with stronger visual perception capabilities will suffer from modality conflict and experience a decline in performance. On the contrary, MLLMs with weaker visual capabilities are insensitive to this conflict, and their performance changes little. 

Toxic Captions. Under the Cap only setting, as revealed in our Case Study~\ref{case_study}, some "toxic" captions can mislead the reasoning model. For example, when using captions generated by Qwen2.5-VL-7B-Instruct in the Vision-Only tasks, some MLLMs exhibit performance degradation. 

In short, by introducing CapGeo-Bench, we build a challenging benchmark that establishes a new standard for the geometric captioning community, encouraging future research to address these limitations.
Beyond serving as an evaluation benchmark, CapGeo-Bench's fine-grained scoring naturally lends itself as a reward signal source for RL —an exciting direction for future work.

\section{Conclusion}
In this paper, we investigated the root cause of the persistent weakness of MLLMs in multimodal geometric reasoning. Using CapGeo as a controlled probe, we empirically verified that the bottleneck lies in visual perception rather than reasoning. Building on this finding, we proposed CapGeo-Bench, a high-quality benchmark of 4{,}641 geometric figure–caption pairs equipped with a fine-grained keypoint-based evaluation metric, designed to systematically assess MLLMs' geometric information extraction capability. Scores on CapGeo-Bench show a strong positive correlation with downstream reasoning performance, establishing it as a powerful tool for evaluating geometric perception capability. Together, CapGeo and CapGeo-Bench pave a new path for future advances in multimodal geometric perception and reasoning.

\section{Limitations}
Due to the influence and limitations of objective factors and actual conditions, despite our best efforts to avoid them, there are still some limitations in this study. Firstly, the limited time and computing resources prevent our experimental framework CapGeo from conducting a comprehensive evaluation on all SOTA models and datasets. Secondly, although CapGeo-Bench has taken strict measures in terms of quality control, such as the screening and training of annotators and the final review system, the inherent inaccuracy in the manual annotation process may still lead to minor mistakes and errors in the dataset.

\bibliography{main}

@article{zhang2024mathverse,
  title={Mathverse: Does your multi-modal llm truly see the diagrams in visual math problems?},
  author={Zhang, Renrui and Jiang, Dongzhi and Zhang, Yichi and Lin, Haokun and Guo, Ziyu and Qiu, Pengshuo and Zhou, Aojun and Lu, Pan and Chang, Kai-Wei and Gao, Peng and others},
  journal={arXiv preprint arXiv:2403.14624},
  year={2024}
}

@article{lu2023mathvista,
  title={Mathvista: Evaluating mathematical reasoning of foundation models in visual contexts},
  author={Lu, Pan and Bansal, Hritik and Xia, Tony and Liu, Jiacheng and Li, Chunyuan and Hajishirzi, Hannaneh and Cheng, Hao and Chang, Kai-Wei and Galley, Michel and Gao, Jianfeng},
  journal={arXiv preprint arXiv:2310.02255},
  year={2023}
}

@inproceedings{llava,
  author       = {Haotian Liu and
                  Chunyuan Li and
                  Qingyang Wu and
                  Yong Jae Lee},
  title        = {Visual Instruction Tuning},
  booktitle    = {Advances in Neural Information Processing Systems 36: Annual Conference
                  on Neural Information Processing Systems 2023, NeurIPS 2023, New Orleans,
                  LA, USA, December 10 - 16, 2023},
  year         = {2023},
}

@article{gao2023g,
  title={G-llava: Solving geometric problem with multi-modal large language model},
  author={Gao, Jiahui and Pi, Renjie and Zhang, Jipeng and Ye, Jiacheng and Zhong, Wanjun and Wang, Yufei and Hong, Lanqing and Han, Jianhua and Xu, Hang and Li, Zhenguo and others},
  journal={arXiv preprint arXiv:2312.11370},
  year={2023}
}

@article{guo2025deepseek,
  title={Deepseek-r1: Incentivizing reasoning capability in llms via reinforcement learning},
  author={Guo, Daya and Yang, Dejian and Zhang, Haowei and Song, Junxiao and Zhang, Ruoyu and Xu, Runxin and Zhu, Qihao and Ma, Shirong and Wang, Peiyi and Bi, Xiao and others},
  journal={arXiv preprint arXiv:2501.12948},
  year={2025}
}

@inproceedings{mathew2021docvqa,
  title={Docvqa: A dataset for vqa on document images},
  author={Mathew, Minesh and Karatzas, Dimosthenis and Jawahar, CV},
  booktitle={Proceedings of the IEEE/CVF winter conference on applications of computer vision},
  pages={2200--2209},
  year={2021}
}

@article{huang2025gemini,
  title={Gemini 2.5 Pro capable of winning gold at IMO 2025},
  author={Huang, Yichen and Yang, Lin F},
  journal={arXiv preprint arXiv:2507.15855},
  year={2025}
}

@article{bai2025qwen2,
  title={Qwen2. 5-vl technical report},
  author={Bai, Shuai and Chen, Keqin and Liu, Xuejing and Wang, Jialin and Ge, Wenbin and Song, Sibo and Dang, Kai and Wang, Peng and Wang, Shijie and Tang, Jun and others},
  journal={arXiv preprint arXiv:2502.13923},
  year={2025}
}

@article{yao2024minicpm,
  title={Minicpm-v: A gpt-4v level mllm on your phone},
  author={Yao, Yuan and Yu, Tianyu and Zhang, Ao and Wang, Chongyi and Cui, Junbo and Zhu, Hongji and Cai, Tianchi and Li, Haoyu and Zhao, Weilin and He, Zhihui and others},
  journal={arXiv preprint arXiv:2408.01800},
  year={2024}
}

@misc{anthropic2025claude,
  author       = {{Anthropic}},
  title        = {Transparency Hub: Model Report},
  howpublished = {\url{https://www.anthropic.com/transparency/model-report}},
  year         = {2025}
}

@article{qiu2025physics,
  title={Physics Supernova: AI Agent Matches Elite Gold Medalists at IPhO 2025},
  author={Qiu, Jiahao and Shi, Jingzhe and Juan, Xinzhe and Zhao, Zelin and Geng, Jiayi and Liu, Shilong and Wang, Hongru and Wu, Sanfeng and Wang, Mengdi},
  journal={arXiv preprint arXiv:2509.01659},
  year={2025}
}

@techreport{gemini2025report,
  title     = {Gemini 2.5: Pushing the Frontier with Advanced Reasoning, Multimodality, Long Context, and Next Generation Agentic Capabilities},
  author    = {{Gemini Team, Google DeepMind}},
  year      = {2025},
  institution = {Google DeepMind},
  url       = {https://storage.googleapis.com/deepmind-media/gemini/gemini_v2_5_report.pdf}
}

@misc{openai2025o3,
  title        = {Introducing OpenAI o3 and o4-mini},
  author       = {OpenAI},
  year         = {2025},
  howpublished = {\url{https://openai.com/index/introducing-o3-and-o4-mini/}},
  note         = {Accessed: 2025-09-28}
}

@article{qwen25vl,
  title={Qwen2. 5-vl technical report},
  author={Bai, Shuai and Chen, Keqin and Liu, Xuejing and Wang, Jialin and Ge, Wenbin and Song, Sibo and Dang, Kai and Wang, Peng and Wang, Shijie and Tang, Jun and others},
  journal={arXiv preprint arXiv:2502.13923},
  year={2025}
}

@article{mmgeolm,
  title={Hard Negative Contrastive Learning for Fine-Grained Geometric Understanding in Large Multimodal Models},
  author={Sun, Kai and Bai, Yushi and Yang, Zhen and Zhang, Jiajie and Qi, Ji and Hou, Lei and Li, Juanzi},
  journal={arXiv preprint arXiv:2505.20152},
  year={2025}
}

@article{mcot1,
  title={Multimodal chain-of-thought reasoning in language models},
  author={Zhang, Zhuosheng and Zhang, Aston and Li, Mu and Zhao, Hai and Karypis, George and Smola, Alex},
  journal={arXiv preprint arXiv:2302.00923},
  year={2023}
}

@inproceedings{mcot2,
  title={Dcot: Dual chain-of-thought prompting for large multimodal models},
  author={Jia, Zixi and Liu, Jiqiang and Li, Hexiao and Liu, Qinghua and Gao, Hongbin},
  booktitle={The 16th Asian Conference on Machine Learning (Conference Track)},
  year={2024}
}

@article{mcot3,
  title={Mc-cot: A modular collaborative cot framework for zero-shot medical-vqa with llm and mllm integration},
  author={Wei, Lai and Wang, Wenkai and Shen, Xiaoyu and Xie, Yu and Fan, Zhihao and Zhang, Xiaojin and Wei, Zhongyu and Chen, Wei},
  journal={arXiv preprint arXiv:2410.04521},
  year={2024}
}

@inproceedings{mcot4,
  title={Cantor: Inspiring multimodal chain-of-thought of mllm},
  author={Gao, Timin and Chen, Peixian and Zhang, Mengdan and Fu, Chaoyou and Shen, Yunhang and Zhang, Yan and Zhang, Shengchuan and Zheng, Xiawu and Sun, Xing and Cao, Liujuan and others},
  booktitle={Proceedings of the 32nd ACM International Conference on Multimedia},
  pages={9096--9105},
  year={2024}
}

@article{mo1-1,
  title={Llama-berry: Pairwise optimization for o1-like olympiad-level mathematical reasoning},
  author={Zhang, Di and Wu, Jianbo and Lei, Jingdi and Che, Tong and Li, Jiatong and Xie, Tong and Huang, Xiaoshui and Zhang, Shufei and Pavone, Marco and Li, Yuqiang and others},
  journal={arXiv preprint arXiv:2410.02884},
  year={2024}
}

@article{mo1-2,
  title={Mulberry: Empowering mllm with o1-like reasoning and reflection via collective monte carlo tree search},
  author={Yao, Huanjin and Huang, Jiaxing and Wu, Wenhao and Zhang, Jingyi and Wang, Yibo and Liu, Shunyu and Wang, Yingjie and Song, Yuxin and Feng, Haocheng and Shen, Li and others},
  journal={arXiv preprint arXiv:2412.18319},
  year={2024}
}

@article{mo1-3,
  title={Llamav-o1: Rethinking step-by-step visual reasoning in llms},
  author={Thawakar, Omkar and Dissanayake, Dinura and More, Ketan and Thawkar, Ritesh and Heakl, Ahmed and Ahsan, Noor and Li, Yuhao and Zumri, Mohammed and Lahoud, Jean and Anwer, Rao Muhammad and others},
  journal={arXiv preprint arXiv:2501.06186},
  year={2025}
}

@article{mr1-1,
  title={Visual-rft: Visual reinforcement fine-tuning},
  author={Liu, Ziyu and Sun, Zeyi and Zang, Yuhang and Dong, Xiaoyi and Cao, Yuhang and Duan, Haodong and Lin, Dahua and Wang, Jiaqi},
  journal={arXiv preprint arXiv:2503.01785},
  year={2025}
}

@article{mr1-2,
  title={Vlm-r1: A stable and generalizable r1-style large vision-language model},
  author={Shen, Haozhan and Liu, Peng and Li, Jingcheng and Fang, Chunxin and Ma, Yibo and Liao, Jiajia and Shen, Qiaoli and Zhang, Zilun and Zhao, Kangjia and Zhang, Qianqian and others},
  journal={arXiv preprint arXiv:2504.07615},
  year={2025}
}

@article{mmsurvey,
  title={Perception, reason, think, and plan: A survey on large multimodal reasoning models},
  author={Li, Yunxin and Liu, Zhenyu and Li, Zitao and Zhang, Xuanyu and Xu, Zhenran and Chen, Xinyu and Shi, Haoyuan and Jiang, Shenyuan and Wang, Xintong and Wang, Jifang and others},
  journal={arXiv preprint arXiv:2505.04921},
  year={2025}
}

@article{mathv,
  title={Measuring multimodal mathematical reasoning with math-vision dataset},
  author={Wang, Ke and Pan, Junting and Shi, Weikang and Lu, Zimu and Ren, Houxing and Zhou, Aojun and Zhan, Mingjie and Li, Hongsheng},
  journal={Advances in Neural Information Processing Systems},
  volume={37},
  pages={95095--95169},
  year={2024}
}

@inproceedings{yang2021textvqa,
  title={Tap: Text-aware pre-training for text-vqa and text-caption},
  author={Yang, Zhengyuan and Lu, Yijuan and Wang, Jianfeng and Yin, Xi and Florencio, Dinei and Wang, Lijuan and Zhang, Cha and Zhang, Lei and Luo, Jiebo},
  booktitle={Proceedings of the IEEE/CVF conference on computer vision and pattern recognition},
  pages={8751--8761},
  year={2021}
}

@article{fu2024ocrbench,
  title={Ocrbench v2: An improved benchmark for evaluating large multimodal models on visual text localization and reasoning},
  author={Fu, Ling and Kuang, Zhebin and Song, Jiajun and Huang, Mingxin and Yang, Biao and Li, Yuzhe and Zhu, Linghao and Luo, Qidi and Wang, Xinyu and Lu, Hao and others},
  journal={arXiv preprint arXiv:2501.00321},
  year={2024}
}

@article{liu2024ocrbench,
  title={Ocrbench: on the hidden mystery of ocr in large multimodal models},
  author={Liu, Yuliang and Li, Zhang and Huang, Mingxin and Yang, Biao and Yu, Wenwen and Li, Chunyuan and Yin, Xu-Cheng and Liu, Cheng-Lin and Jin, Lianwen and Bai, Xiang},
  journal={Science China Information Sciences},
  volume={67},
  number={12},
  pages={220102},
  year={2024},
  publisher={Springer}
}

@article{ocrreasoning,
  title={Ocr-reasoning benchmark: Unveiling the true capabilities of mllms in complex text-rich image reasoning},
  author={Huang, Mingxin and Shi, Yongxin and Peng, Dezhi and Lai, Songxuan and Xie, Zecheng and Jin, Lianwen},
  journal={arXiv preprint arXiv:2505.17163},
  year={2025}
}

@article{reasoningocr,
  title={Reasoning-OCR: Can Large Multimodal Models Solve Complex Logical Reasoning Problems from OCR Cues?},
  author={He, Haibin and Ye, Maoyuan and Zhang, Jing and Cai, Xiantao and Liu, Juhua and Du, Bo and Tao, Dacheng},
  journal={arXiv preprint arXiv:2505.12766},
  year={2025}
}

@article{captioneval,
  title={Benchmarking and improving detail image caption},
  author={Dong, Hongyuan and Li, Jiawen and Wu, Bohong and Wang, Jiacong and Zhang, Yuan and Guo, Haoyuan},
  journal={arXiv preprint arXiv:2405.19092},
  year={2024}
}

@article{captioneval2,
  title={Image captioning evaluation in the age of multimodal llms: Challenges and future perspectives},
  author={Sarto, Sara and Cornia, Marcella and Cucchiara, Rita},
  journal={arXiv preprint arXiv:2503.14604},
  year={2025}
}

@inproceedings{cider,
  title={Cider: Consensus-based image description evaluation},
  author={Vedantam, Ramakrishna and Lawrence Zitnick, C and Parikh, Devi},
  booktitle={Proceedings of the IEEE conference on computer vision and pattern recognition},
  pages={4566--4575},
  year={2015}
}

@inproceedings{anderson2016spice,
  title={Spice: Semantic propositional image caption evaluation},
  author={Anderson, Peter and Fernando, Basura and Johnson, Mark and Gould, Stephen},
  booktitle={European conference on computer vision},
  pages={382--398},
  year={2016},
  organization={Springer}
}

@article{chen2021geoqa,
  title={Geoqa: A geometric question answering benchmark towards multimodal numerical reasoning},
  author={Chen, Jiaqi and Tang, Jianheng and Qin, Jinghui and Liang, Xiaodan and Liu, Lingbo and Xing, Eric P and Lin, Liang},
  journal={arXiv preprint arXiv:2105.14517},
  year={2021}
}

@article{lu2021inter,
  title={Inter-gps: Interpretable geometry problem solving with formal language and symbolic reasoning},
  author={Lu, Pan and Gong, Ran and Jiang, Shibiao and Qiu, Liang and Huang, Siyuan and Liang, Xiaodan and Zhu, Song-Chun},
  journal={arXiv preprint arXiv:2105.04165},
  year={2021}
}

@inproceedings{caption1,
  title={Differential-perceptive and retrieval-augmented mllm for change captioning},
  author={Zhang, Xian and Wen, Haokun and Wu, Jianlong and Qin, Pengda and Xue', Hui and Nie, Liqiang},
  booktitle={Proceedings of the 32nd ACM International Conference on Multimedia},
  pages={4148--4157},
  year={2024}
}

@article{caption2,
  title={Toward robust hyper-detailed image captioning: A multiagent approach and dual evaluation metrics for factuality and coverage},
  author={Lee, Saehyung and Yoon, Seunghyun and Bui, Trung and Shi, Jing and Yoon, Sungroh},
  journal={arXiv preprint arXiv:2412.15484},
  year={2024}
}

@inproceedings{caption3,
  title={Dreamlip: Language-image pre-training with long captions},
  author={Zheng, Kecheng and Zhang, Yifei and Wu, Wei and Lu, Fan and Ma, Shuailei and Jin, Xin and Chen, Wei and Shen, Yujun},
  booktitle={European Conference on Computer Vision},
  pages={73--90},
  year={2024},
  organization={Springer}
}

@article{zhao2024doclayout,
  title={Doclayout-yolo: Enhancing document layout analysis through diverse synthetic data and global-to-local adaptive perception},
  author={Zhao, Zhiyuan and Kang, Hengrui and Wang, Bin and He, Conghui},
  journal={arXiv preprint arXiv:2410.12628},
  year={2024}
}

@inproceedings{kwon2023efficient,
  title={Efficient Memory Management for Large Language Model Serving with PagedAttention},
  author={Woosuk Kwon and Zhuohan Li and Siyuan Zhuang and Ying Sheng and Lianmin Zheng and Cody Hao Yu and Joseph E. Gonzalez and Hao Zhang and Ion Stoica},
  booktitle={Proceedings of the ACM SIGOPS 29th Symposium on Operating Systems Principles},
  year={2023}
}

@article{jia2024describe,
  title={Describe-then-reason: Improving multimodal mathematical reasoning through visual comprehension training},
  author={Jia, Mengzhao and Zhang, Zhihan and Yu, Wenhao and Jiao, Fangkai and Jiang, Meng},
  journal={arXiv preprint arXiv:2404.14604},
  year={2024}
}

@article{wei2024slow,
  title={Slow Perception: Let's Perceive Geometric Figures Step-by-step},
  author={Wei, Haoran and Yin, Youyang and Li, Yumeng and Wang, Jia and Zhao, Liang and Sun, Jianjian and Ge, Zheng and Zhang, Xiangyu and Jiang, Daxin},
  journal={arXiv preprint arXiv:2412.20631},
  year={2024}
}

@article{chen2025geopqa,
  title={Geopqa: Bridging the visual perception gap in mllms for geometric reasoning},
  author={Chen, Guizhen and Xu, Weiwen and Zhang, Hao and Chan, Hou Pong and Zhao, Deli and Luu, Anh Tuan and Rong, Yu},
  journal={ArXiv, abs/2509.17437, 2025a. URL https://arxiv. org/abs/2509.17437},
  year={2025}
}

@article{peng2024multimath,
  title={Multimath: Bridging visual and mathematical reasoning for large language models},
  author={Peng, Shuai and Fu, Di and Gao, Liangcai and Zhong, Xiuqin and Fu, Hongguang and Tang, Zhi},
  journal={arXiv preprint arXiv:2409.00147},
  year={2024}
}

@article{xing2025caprl,
  title={Caprl: Stimulating dense image caption capabilities via reinforcement learning},
  author={Xing, Long and Dong, Xiaoyi and Zang, Yuhang and Cao, Yuhang and Liang, Jianze and Huang, Qidong and Wang, Jiaqi and Wu, Feng and Lin, Dahua},
  journal={arXiv preprint arXiv:2509.22647},
  year={2025}
}

@article{xiao2025perception,
  title={Perception-R1: Advancing Multimodal Reasoning Capabilities of MLLMs via Visual Perception Reward},
  author={Xiao, Tong and Xu, Xin and Huang, Zhenya and Gao, Hongyu and Liu, Quan and Liu, Qi and Chen, Enhong},
  journal={arXiv preprint arXiv:2506.07218},
  year={2025}
}

@inproceedings{10.1145/3774904.3792075,
author = {Zheng, Leqi and Zhang, Jiajun and Chen, Canzhi and Wang, Chaokun and Li, Hongwei and Li, Yuying and Mao, Yaoxin and Yan, Shannan and Song, Zixin and Feng, Zhiyuan and Kang, Zhaolu and Chen, Zirong and Zhang, Hang and Liu, Qiang and Wang, Liang and Liu, Ziyang},
title = {What Should I Cite? A RAG Benchmark for Academic Citation Prediction},
year = {2026},
isbn = {9798400723070},
publisher = {Association for Computing Machinery},
address = {New York, NY, USA},
url = {https://doi.org/10.1145/3774904.3792075},
doi = {10.1145/3774904.3792075},
booktitle = {Proceedings of the ACM Web Conference 2026},
pages = {1852-1863},
numpages = {12},
location = {United Arab Emirates},
series = {WWW '26}
}

@article{xia2025visionary,
  title={Visionary-r1: Mitigating shortcuts in visual reasoning with reinforcement learning},
  author={Xia, Jiaer and Zang, Yuhang and Gao, Peng and Li, Sharon and Zhou, Kaiyang},
  journal={arXiv preprint arXiv:2505.14677},
  year={2025}
}

@article{chen2025perception,
  title={Perception before reasoning: Two-stage reinforcement learning for visual reasoning in vision-language models},
  author={Chen, Yan and Li, Long and Xi, Teng and Zeng, Long and Wang, Jingdong},
  journal={arXiv preprint arXiv:2509.13031},
  year={2025}
}
\clearpage
\appendix 

\section{Prompt Appendix}\label{Appendix.prompt}

\begin{figure*}
    \centering
    \includegraphics[width=1\linewidth]{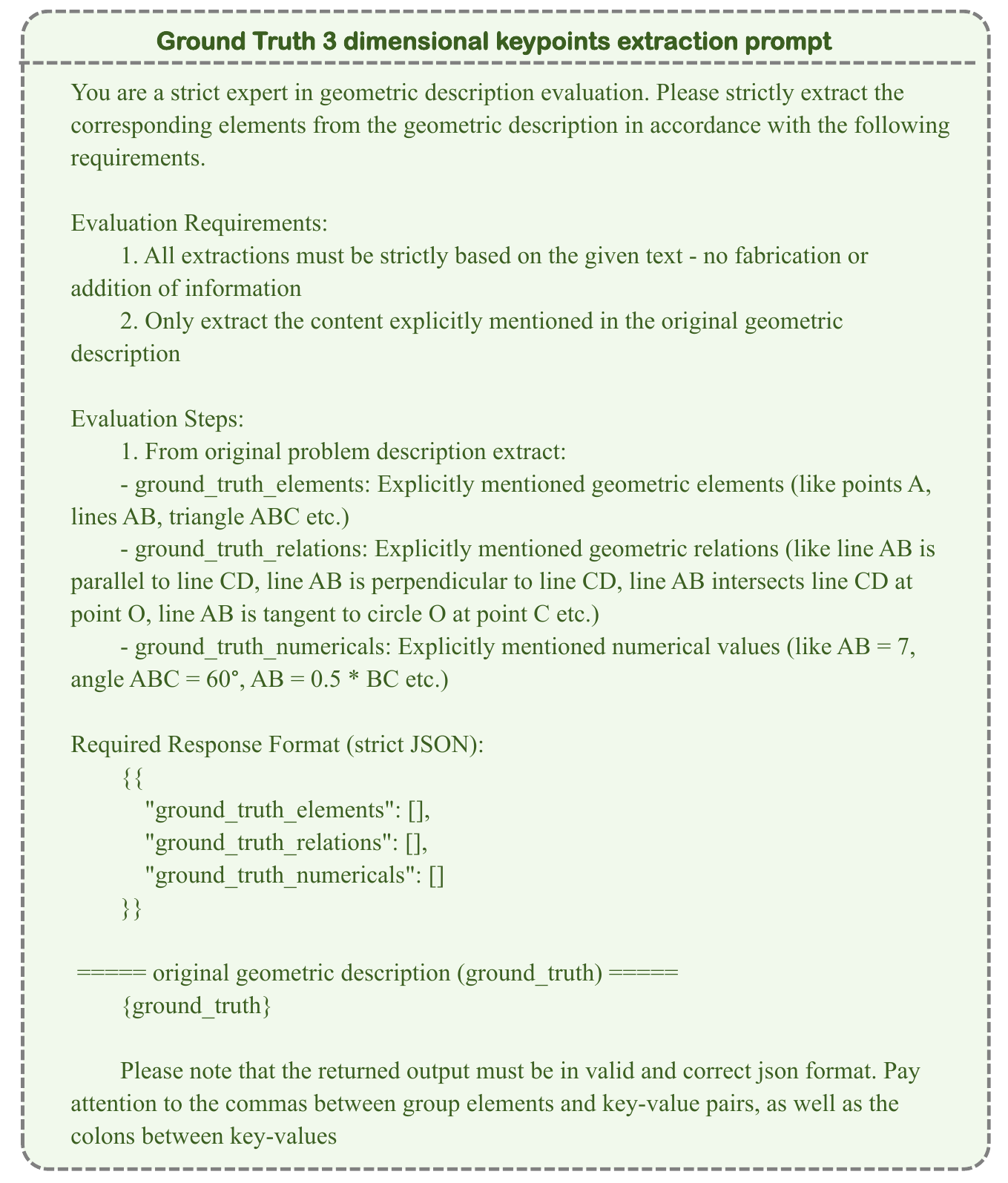}
    \caption{Ground Truth 3 dimensional keypoints set extraction prompt for LLM during CapGeo-Bench evaluation}
    \label{fig:prompt1}
\end{figure*}

\begin{figure*}
    \centering
    \includegraphics[width=1\linewidth]{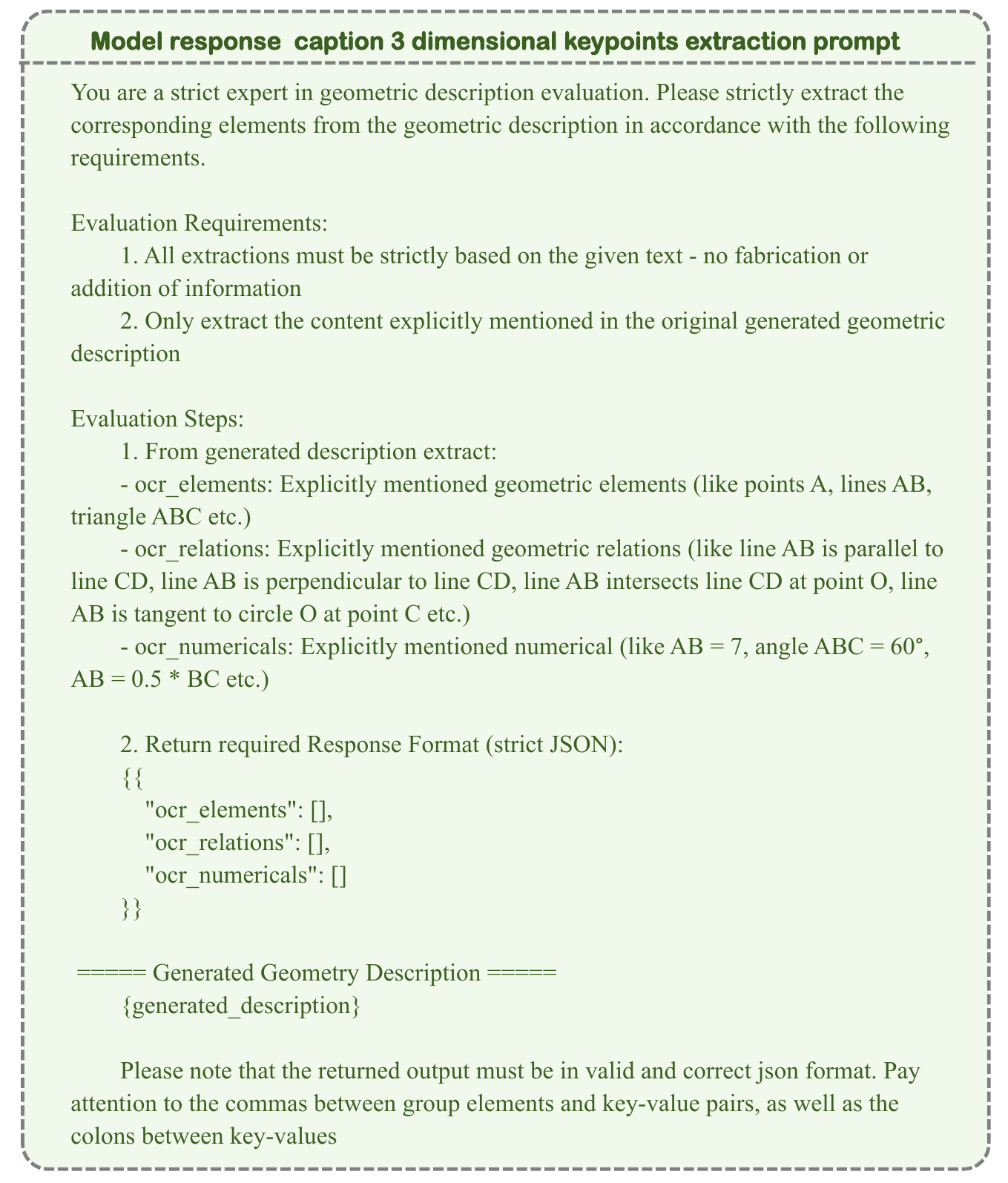}
    \caption{Model Response Caption 3 dimensional keypoints set extraction prompt for LLM during CapGeo-Bench evaluation}
    \label{fig:prompt2}
\end{figure*}

\begin{figure*}
    \centering
    \includegraphics[width=1.0\linewidth]{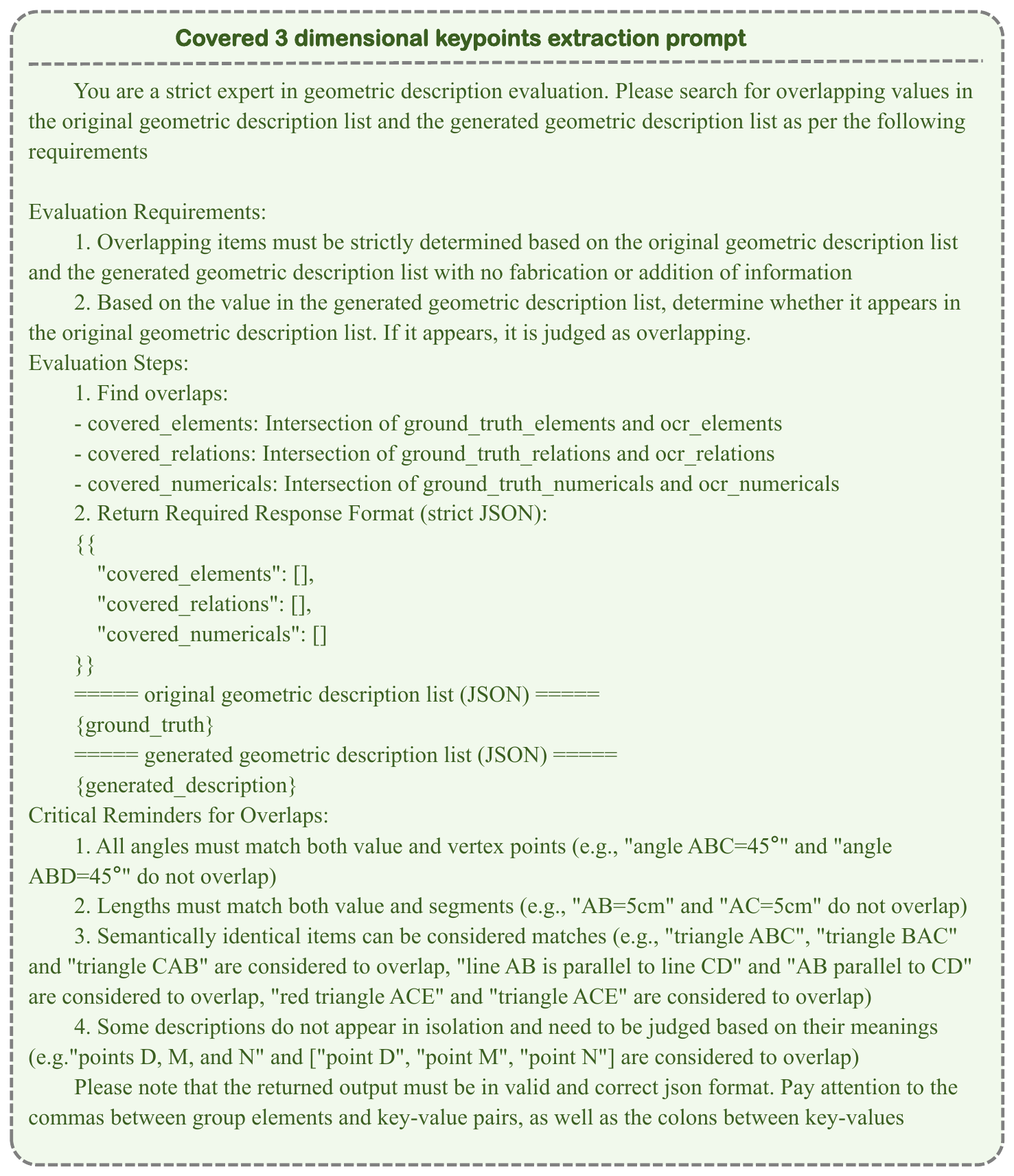}
    \caption{Covered 3 dimensional keypoints extraction prompt for LLM during CapGeo-Bench evaluation}
    \label{fig:prompt3}
\end{figure*}

In this section, we have detailedly listed all the prompts used in the experiments of the articles. In CapGeo, prompts related to captioning and reasoning are involved in Figure~\ref{fig:prompt1}. In CapGeo-Bench, the prompt related to captioning is involved in Figure~\ref{fig:prompt1}, the prompt for keyword extraction during the evaluation process is involved in Figure~\ref{fig:prompt2}, and the prompt for covered keypoints extraction is involved in Figure~\ref{fig:prompt3}. All prompts have been carefully and scientifically designed and iterated through experiments, and have been guided and verified by domain experts, resulting in excellent final effects.

\section{Full CapGeo Results with Additional Captioners}\label{Appendix.fullcap}
For brevity, the main paper (Table~\ref{tab:capgeo_preview}) reports CapGeo results using GPT-o3 as the captioner, since it produces the strongest captions and thus most clearly exposes the perception bottleneck. Here we provide the full results with three additional captioners (GPT-4o, Qwen2.5-VL-72B-Instruct, Qwen2.5-VL-7B-Instruct) for completeness in Table~\ref{tab:full_cap}. The trends across all captioners are consistent with the conclusions in the main text: stronger captioners yield larger reasoning gains, while weaker captioners can introduce modality conflict and toxic captions.

\begin{table*}[htbp]
\renewcommand{\arraystretch}{1.2}
\small
\centering
\begin{tabular}{
    p{3.4cm}|
    >{\centering\arraybackslash}p{0.8cm}
    *{4}{>{\centering\arraybackslash}p{0.6cm}>{\centering\arraybackslash}p{1.3cm}}
}
\toprule
\multirow{2}{*}{\diagbox[width=3.8cm, height=3.8em]{\makecell{\textbf{Reasoning}\\\textbf{Model}}}{\makecell{\textbf{Captioning}\\\textbf{Model}}} }
&\textbf{w/o}
& \multicolumn{2}{c}{\textbf{GPT-o3}}
& \multicolumn{2}{c}{\textbf{GPT-4o}}
& \multicolumn{2}{c}{\textbf{\makecell{Qwen2.5-VL\\ -72B-Instruct}}}
& \multicolumn{2}{c}{\textbf{\makecell{Qwen2.5-VL\\ -7B-Instruct}}} \\
\cmidrule(lr){2-2}\cmidrule(lr){3-4} \cmidrule(lr){5-6} \cmidrule(lr){7-8} \cmidrule(lr){9-10}
& \textbf{Img}
& \textbf{Cap} & \textbf{Cap+Img}
& \textbf{Cap} & \textbf{Cap+Img}
& \textbf{Cap} & \textbf{Cap+Img}
& \textbf{Cap} & \textbf{Cap+Img} \\
\midrule

\multicolumn{10}{c}{\textbf{Vision Only}} \\
\midrule
GPT-o3                 & 74.6 & 60.3 & \textbf{75.8} & 54.6 & 67.8 & 60.3 & 64.6 & 28.7 & 37.3 \\  
Gemini-2.5-pro         & 66.4 & 60.3 & \textbf{73.6} & 57.1 & 56.6 & 62.2 & 60.5 & 37.2 & 37.8 \\  
Claude-Opus-4-20250514 & 44.8 & 59.4 & \textbf{73.0} & 51.8 & 48.5 & 60.9 & 57.9 & 29.7 & 36.7 \\  
Qwen2.5-VL-72B-Instruct         & 8.6  & 66.1 & \textbf{66.4} & 51.1 & 59.0 & 57.0 & 56.6 & 33.6 & 36.4 \\  
Qwen2.5-VL-7B-Instruct          & 7.1  & \textbf{61.9} & 60.0 & 45.9 & 51.5 & 52.7 & 49.4 & 31.0 & 35.2 \\  
\midrule
DeepSeek-R1            & --   & \textbf{68.2} & --   & 57.1 & --   & 62.3 & --   & 36.5 & --   \\  
Qwen2.5-72B-Instruct   & --   & \textbf{66.1} & --   & 50.6 & --   & 54.8 & --   & 31.2 & --   \\  
Qwen2.5-7B-Instruct    & --   & \textbf{63.3} & --   & 47.1 & --   & 53.4 & --   & 31.5 & --   \\  
\midrule

\multicolumn{10}{c}{\textbf{Vision Intensive}} \\
\midrule
GPT-o3                 & 73.0 & \textbf{82.5} & 78.3 & 76.3 & 81.3 & 76.3 & 60.0 & 75.4 & 46.6 \\  
Gemini-2.5-pro         & 79.6 & 83.0 & \textbf{85.4} & 76.1 & 78.0 & 77.9 & 59.0 & 75.1 & 47.3 \\ 
Claude-Opus-4-20250514 & 61.8 & 79.3 & \textbf{85.5} & 75.0 & 76.7 & 74.1 & 68.9 & 74.1 & 47.5 \\   
Qwen2.5-VL-72B-Instruct         & 58.0 & \textbf{74.9} & 72.6 & 66.5 & 64.8 & 70.3 & 71.6 & 62.6 & 62.7 \\  
Qwen2.5-VL-7B-Instruct          & 43.3 & 63.2 & 60.3 & 61.5 & 60.8 & \textbf{67.4} & 70.2 & 60.8 & 62.7 \\  
\midrule
DeepSeek-R1            & --   & 71.9 & --   & 73.9 & --   & \textbf{75.3} & --   & 70.1 & --   \\  
Qwen2.5-72B-Instruct   & --   & \textbf{73.2} & --   & 65.6 & --   & 69.7 & --   & 62.3 & --   \\  
Qwen2.5-7B-Instruct    & --   & 65.9 & --   & 60.2 & --   & \textbf{68.3} & --   & 62.3 & --   \\  
\bottomrule
\end{tabular}
\caption{Full CapGeo results on MathVerse with all four captioners across three input modalities.}
\label{tab:full_cap}
\end{table*}

\section{Full Results on MathVista and GeoQA}\label{Appendix.mvgeoqa}

For brevity, the main paper focuses on MathVerse to verify the perception bottleneck. We additionally report CapGeo results on MathVista and GeoQA in Table~\ref{tab:mv_geoqa_combined}. Trends are consistent with the conclusions in the main text: stronger captioners yield substantial reasoning gains, and the gap between Image-only and Caption-assisted settings reconfirms the perception bottleneck.

\begin{table*}[t]
\centering
\resizebox{\linewidth}{!}{
\begin{tabular}{l|ccccc|ccccc}
\toprule
\multirow{2}{*}{\textbf{Model}} 
& \multicolumn{5}{c|}{\textbf{MathVista}} 
& \multicolumn{5}{c}{\textbf{GeoQA}} \\
\cmidrule(lr){2-6}\cmidrule(lr){7-11}
& \textbf{-} & \textbf{GPT-o3} & \textbf{GPT-4o} & \textbf{Qwen-72B} & \textbf{Qwen-7B}
& \textbf{-} & \textbf{GPT-o3} & \textbf{GPT-4o} & \textbf{Qwen-72B} & \textbf{Qwen-7B} \\
\midrule
Claude-Opus-4-20250514       & 76.1 & 89.4 & 65.7 & \textbf{91.7} & 88.4 & 57.0 & 92.0 & 91.0 & \textbf{92.5} & 91.5 \\
Gemini-2.5-pro-preview-05-06 & 94.9 & 96.3 & 95.8 & \textbf{98.1} & 97.7 & 90.0 & 95.0 & 96.5 & 97.0 & \textbf{97.5} \\
GPT-4o                   & 61.6 & \textbf{65.7} & 62.0 & 63.9 & 62.5 & 61.0 & \textbf{63.0} & \textbf{63.0} & 62.5 & 57.5 \\
GPT-o3                   & 93.1 & 94.9 & 94.0 & 94.4 & \textbf{95.8} & 92.5 & 94.5 & 94.5 & 93.5 & \textbf{95.0} \\
Qwen2.5-VL-72B-Instruct      & 75.9 & \textbf{92.6} & 91.7 & 92.1 & 70.4 & 68.5 & \textbf{73.5} & 72.0 & 70.0 & \textbf{73.5} \\
\bottomrule
\end{tabular}}
\caption{Comparison of MLLMs on \textbf{MathVista} and \textbf{GeoQA} benchmarks with different captioning models.}
\label{tab:mv_geoqa_combined}
\end{table*}

\section{Ablation Study on Caption Style}\label{Appendix.style}

\begin{figure*}
    \centering
    \includegraphics[width=1\linewidth]{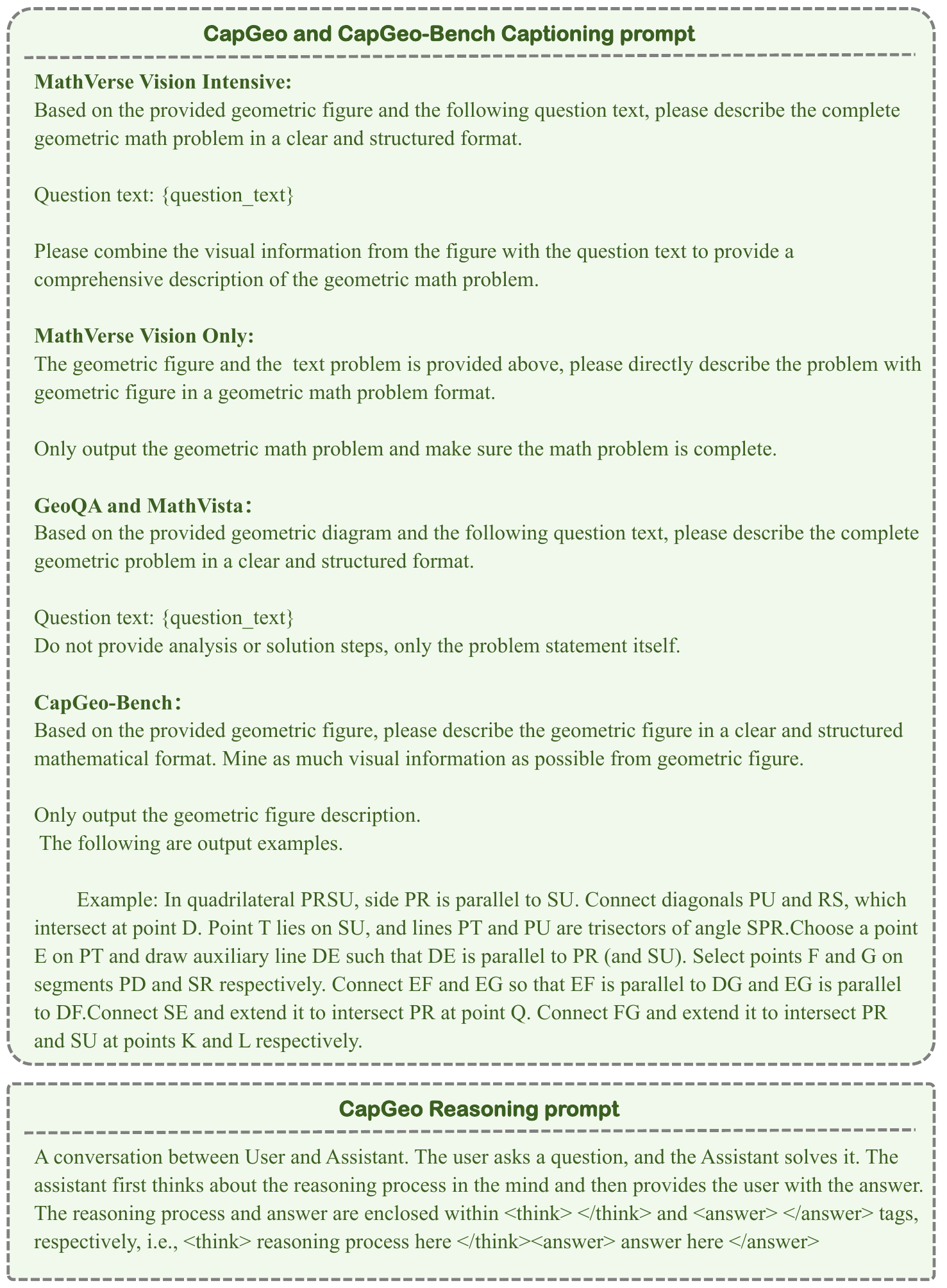}
    \caption{CapGeo and CapGeo-Bench captioning prompt for Captioner Model and reasoning prompt for Reasoning Models during the experiments}
    \label{fig:caption}
\end{figure*}

A natural concern is whether the gains of CapGeo stem from a specific captioning template rather than from genuine visual information extraction. To address this, we ablate two distinct caption styles while keeping all other factors fixed:

\begin{itemize}
\item \textbf{Natural Problem Reformulation} (used in our main experiments): the captioner is simply asked to ``describe the complete geometric math problem in a clear and structured format'', without enforcing any rigid schema. The output naturally takes the form of a standard geometry problem statement (see Figure~\ref{fig:caption}).

\item \textbf{Explicit Structured Decomposition}: the captioner is forced to output a symbolic JSON schema that explicitly enumerates three fields—\textit{Geometric Elements}, \textit{Element Relationships}, and \textit{Numerical Information}.
\end{itemize}

We re-run CapGeo with both caption styles and report results in Table~\ref{tab:caption_style}.

We observe a consistent degradation when using the highly structured JSON captions across all four reasoning models. We attribute this to two factors: (i) overly formalized JSON representations disrupt the semantic continuity that (M)LLMs naturally exploit during step-by-step geometric reasoning; and (ii) rigid schemas can fragment relational information that is more naturally expressed in mathematical prose. This ablation provides two important takeaways: (1) the gains observed in CapGeo arise from the \textit{accurate extraction of visual information}, not from any specific captioning template; (2) given the inherently formal nature of geometry, asking the captioner for a natural mathematical-style description is itself a mild and well-motivated requirement, not a hand-crafted trick.

\section{Difficulty Classification}\label{Appendix.difficulty}
During the annotation process, the classification of difficulty levels was determined by \textbf{the complexity of geometric diagrams}, which aligns closely with the core motivation of CapGeo-Bench: to evaluate the capability of extracting geometric information. 

Specifically, after reviewing all pre-screened geometric diagrams, we categorized them into four difficulty levels—\textbf{Simple (T1), Moderate (T2), Complex (T3), and Challenging (T4)}—based on the intricacy of their elements and structural composition. Representative diagrams for each difficulty level were provided to annotators to facilitate consistent and accurate labeling.

\section{Annotation Instructions}\label{Appendix.instruction}
To ensure the reliability and accuracy of the annotated data, we adopted a rigorous selection and training process for annotators. Specifically, we recruited candidates from top universities and ranked them based on their high school mathematics performance and scores on geometry reasoning tests, ensuring sufficient mathematical reasoning and geometric spatial understanding capabilities.

Before undertaking the actual annotation tasks, candidates went through two rounds of training and pre-annotation. In the first stage, we provided comprehensive instruction covering the annotation objectives, guidelines, workflow, and quality standards, illustrated with detailed examples. Candidates were then required to perform pre-annotations on 10 geometric figures, which were reviewed by expert examiners. The examiners identified both individual and common issues, which were subsequently addressed in a second training session. Afterward, candidates completed another round of pre-annotation and review. Only those who successfully passed both evaluations were qualified as final annotators.

During the annotation process, annotators were required to follow a rigorous \textbf{Workflow}. 
\begin{itemize}
\item First, they assessed the quality and content of the image to determine whether it was a valid geometric figure; images that did not meet the criteria were discarded. 

\item Next, annotators evaluated the type and difficulty level of the geometric figure. 

\item Finally, annotators carefully examined the construction logic, spatial relations, and numerical information of the figure, and produced a logically structured caption that covered all information in the image without introducing subjective hallucinations. 

\item For highly complex figures or when necessary, annotators were instructed to consult the original descriptions in the source PDF to ensure the accuracy and reliability of the final annotations.
\end{itemize}

Our \textbf{Annotation instructions} are as follows:

\begin{itemize}
    \item \textbf{Strict adherence to expert guidelines}: Annotators were required to strictly adhere to unified guidelines designed by mathematics experts.
    
    \item \textbf{Quality control of visual materials}: Redundant, undescribable, or low-quality images were to be discarded.
    
    \item \textbf{Visual grounding of captions}: All captions had to be strictly grounded in the visual content of the figure, without introducing subjective interpretation or speculation.
    
    \item \textbf{Consistent geometric terminology}: Geometric terminology was required to remain consistent, ensuring standardization and reproducibility across annotations.
    
    \item \textbf{Beyond surface-level description}: Captions were expected to go beyond surface-level elements and capture essential geometric attributes or spatial relations (e.g., for a midpoint, not merely stating that a point lies on a line, but explicitly indicating its geometric significance).
    
    \item \textbf{Hierarchical structure}: Annotations were required to follow a hierarchical structure, prioritizing information according to geometric construction logic rather than presenting items indiscriminately.
    
    \item \textbf{Fine-grained detail annotation}: Fine-grained details such as dashed lines, colored segments, and area representations were also mandated to be annotated, ensuring comprehensive and precise descriptions.
\end{itemize}

\textbf{Annotation Principles and Ground Truth Definition.}
In our study, the geometric elements constituting the Ground Truth are strictly derived from their presence in the caption descriptions. To ensure objectivity, the annotation process precludes subjective interpretation of the diagrams. Instead, annotators are instructed to adhere to the \textbf{logical construction} of the geometric diagrams as described in the original source textbooks. This ensures that all textual descriptions are firmly grounded in the visual information provided by the images.

\textbf{Selection Criteria for Geometric Elements.}
Based on the construction logic, we categorize labeled elements into two types: (1) \textbf{Constructive Elements}, which are directly involved in the geometric construction, and (2) \textbf{Derivable Elements}, which can be inferred from existing logic. \textbf{Only the former are required to be labeled.} The specific protocols are as follows:

\begin{itemize}
    \item \textbf{Explicitly Marked Elements:} Any element that is explicitly marked in the diagram or plays a direct role in the geometric construction sequence must be explicitly mentioned in the caption.
    
    \item \textbf{Implicit but Essential Elements:} If an unmarked point is essential to the subsequent construction logic (e.g., the intersection point of extended segments $AB$ and $CD$, used as a reference for further construction), it must be described in the caption (e.g., denoted as ``the intersection point of $AB$ and $CD$'').
    
    \item \textbf{Incidental Intersections:} Points that result merely from the intersection of existing segments without participating in the logic of geometric construction, and are not explicitly marked, are excluded from annotation.
\end{itemize}

\textbf{Protocol for Spatial Relationships}
Given the interdependence of geometric relationships, our annotation principle strictly follows the original textbook's logical framework rather than providing an exhaustive enumeration of all deducible information:

\begin{itemize}
    \item \textbf{Explicit Definition:} If the textbook explicitly states a relationship (e.g., ``$A$ is the midpoint of $B C$'') and utilizes this for construction, it is included in the caption.
    \item \textbf{Logical Deduction:} If the textbook describes a condition that implies a relationship but does not explicitly state it (e.g., ``$A D$ is the median of $B C$ with intersection point $A$''), we annotate the median relationship but do not require labeling $A$ as the midpoint, even though it is logically deducible.
\end{itemize}

\textbf{Additional Annotation Details}

\begin{itemize}
\item \textbf{For Annotators}, we initially recruited 100 candidates with STEM backgrounds from top universities, and after two rounds of training and pre-annotation evaluations, we selected 24 formal annotators, with an additional 3 domain experts forming an independent review team.

\item \textbf{For Data Filtering and Re-annotation Rates}, after automatic filtering by DocLayout-YOLO, we ultimately retained 4,641 images from the original 5,000, yielding a collection pass rate of 92.8\%. In the formal annotation stage, approximately 18.3\% of the first-round annotations were returned for re-annotation due to missing elements or imprecise relation descriptions. For pre-adjudication agreement, during the pre-annotation stage, Fleiss‘ Kappa across the three dimensions of elements, relations, and numericals were 0.43, 0.54, and 0.62 before training, and improved to 0.78, 0.81, and 0.86 after systematic training, further reaching 0.83, 0.87, and 0.91 among the 24 selected formal annotators.

\item \textbf{For the Bilingual Process}, the Chinese annotations were reviewed by experts, then independently translated into English by two professional translators, and finally cross-checked by an expert with overseas experience to verify consistency and resolve any disagreements, ensuring terminological uniformity.
\end{itemize}

\textbf{Examination.} To further ensure the quality and consistency of the annotations, we implemented a rigorous cross-review mechanism. All annotation results were independently examined by a review panel consisting of three expert evaluators. For samples that failed the initial review, re-annotations were conducted by different members of the annotation team, and the revised results were resubmitted for review. This iterative process of cross-annotation and review effectively ensured both accuracy and consistency of the final annotations.

We adhere to labour ethics during the whole annotation process, ensuring all annotators were well-paid, with each receiving 10 dollars per geometric figure annotation. 

\section{Spatial Relations}\label{Appendix.Spatial}
To comprehensively extract the spatial relations in the caption, we have classified the following spatial relationship classification system based on the composition of the relationship subjects. We believe that the following contents all fall within the scope of spatial relations. This system encompasses five core types of relationships:

(1) The Point-Line Relations describe the positional dependence between points and lines: Midpoint, Foot of the perpendicular, Intersection point, Trisection point, Endpoint;

(2) The Line-Line Relations depict the relationship between line elements: Perpendicular, Parallel, Oblique intersection, Coincidence;

(3) The Line-Shape Relations characterize the structural connection between a straight line and complex geometric shapes: Angle bisector, Diagonal, Median, Altitude, Chord, Tangent, Diameter;

(4) The Point-Shape Relations define the characteristic position of a point relative to a complex figure: Centroid, Orthocenter, Circumcenter, Incenter, Vertex, Circle center;

(5) The Shape-Shape Relations describe the macroscopic spatial configuration among complete geometric figures: Disjoint, Tangency, Intersection, Containment, Congruence, Similarity, Concentric, Inscribed, Circumscribed.

\section{Correlation Analysis Details}\label{app:correlation}

\subsection{Definition and Sample Size}

We used the Pearson correlation coefficient ($r$) to quantify the relationship between CapGeo-Bench scores and downstream geometric reasoning performance.

For the macro-level analysis aggregating all evaluated models, the sample size is $N = 32$ for Cap Only mode (including both MLLMs and pure-text LLMs) and $N = 20$ for Cap+Img mode (MLLMs only, as pure-text LLMs cannot process image inputs). All reported $p$-values are two-tailed.

\subsection{Correlation Results by Mode}

Under the Vision Only setting, Cap Only yields an average correlation of $r = 0.900$ ($p = 3.87 \times 10^{-11}$, highly significant), while Cap+Img yields $r = 0.953$ ($p = 1.22 \times 10^{-7}$, highly significant).

Under the Vision Intensive setting, Cap Only gives $r = 0.708$ ($p = 4.21 \times 10^{-2}$, significant), and Cap+Img gives $r = 0.633$ ($p = 9.85 \times 10^{-4}$, highly significant).

\subsection{Leave-One-Out Robustness Analysis}

To verify that these strong correlations are not dominated by a few extreme captioners, we conducted a leave-one-out (LOO) analysis on the core Vision Only (Cap Only) setting.

After removing the highest-scoring captioner (GPT-o3), the correlation remains exceptionally high at $r = 0.983$ ($p = 1.64 \times 10^{-11}$). After removing the lowest-scoring captioner (Qwen2.5-VL-7B), the correlation remains robust at $r = 0.726$ ($p = 2.17 \times 10^{-4}$).

These results demonstrate that our correlation is exceptionally stable and effectively rules out the possibility of being dominated by outliers.

\section{Case study}\label{Appendix.Case}

The following case is derived from image\_73.png in the MathVerse dataset:

\begin{tcolorbox}[
    title={\textbf{toxic caption case}},
    label={case_study},
    breakable,
    colback=blue!5!white,
    colframe=blue!75!black,
    fonttitle=\bfseries,
    colbacktitle=blue!85!black,
    enhanced,
    drop shadow={black!50!white},
    attach boxed title to top left={xshift=5mm, yshift=-2mm},
    boxed title style={size=small, colframe=blue!85!black}
]
\textbf{Correct Example (GPT-o3)}

Caption ($\surd$): 

\quad Points $A$, $B$, $C$ lie on circle with center $O$. $\angle ABC = 70^\circ$. Find central angle $\angle AOC$.

Options: 

\quad A) $140^\circ$ B) $130^\circ$ C) $120^\circ$ D) $110^\circ$

Reasoning ($\surd$): 

\quad Central angle = $2 \times$ inscribed angle

Calculation: 

\quad $\angle AOC = 2 \times 70^\circ = 140^\circ$

\textbf{Answer}: A) $140^\circ$

\vspace{0.5em}

\textbf{Error Case (GPT-4o)}

Caption ($\times$): 

\quad \textcolor{red}{Incorrect: "$\angle BOC = 70^\circ$" (should be $\angle ABC = 70^\circ$)}

Misled Reasoning ($\times$): 

\quad Based on wrong caption, calculates:
$180^\circ + 70^\circ + \angle AOC = 360^\circ \implies \angle AOC = 110^\circ$

\textbf{Answer}:D) $110^\circ$
\end{tcolorbox}

\end{document}